\documentclass{article}

\usepackage{microtype}
\usepackage{graphicx}
\usepackage{subcaption}
\usepackage{booktabs} 

\usepackage{hyperref}

\usepackage[preprint]{icml2026}

\usepackage{amsmath}
\usepackage{amssymb}
\usepackage{mathtools}
\usepackage{amsthm}

\usepackage{algorithm}
\usepackage{algorithmic}
\usepackage{amsmath}
\usepackage[table]{xcolor}
\usepackage{booktabs}
\usepackage{multirow}
\usepackage{caption}
\usepackage{enumitem}

\usepackage{hyperref}
\usepackage{nameref}

\usepackage[capitalize,noabbrev]{cleveref}

\theoremstyle{plain}

\theoremstyle{definition}

\theoremstyle{remark}

\usepackage[textsize=tiny]{todonotes}
\definecolor{darkgreen}{rgb}{0.0, 0.5, 0.0}
\icmltitlerunning{CAST: Mitigating Object Hallucination in Large Vision-Language Models via Caption-Guided Visual Attention Steering}

\begin{document}

\twocolumn[
  \icmltitle{CAST: Mitigating Object Hallucination in Large Vision-Language Models \\ via Caption-Guided Visual Attention Steering}



  \icmlsetsymbol{equal}{*}

  \begin{icmlauthorlist}
    \icmlauthor{Qiming Li}{1,equal}
    \icmlauthor{Zekai Ye}{1,equal}
    \icmlauthor{Xiaocheng Feng}{1,2}
    \icmlauthor{Weihong Zhong}{1}
    \icmlauthor{Libo Qin}{3}
    \icmlauthor{Ruihan Chen}{1}
    \icmlauthor{Lei Huang}{1}
    \icmlauthor{Baohang Li}{1}
    \icmlauthor{Kui Jiang}{1}
    \icmlauthor{Yaowei Wang}{1}
    \icmlauthor{Ting Liu}{1}
    \icmlauthor{Bing Qin}{1}
  \end{icmlauthorlist}

  \icmlaffiliation{1}{Harbin Institute of Technology}
  \icmlaffiliation{2}{Peng Cheng Laboratory}
  \icmlaffiliation{3}{Harbin Institute of Technology, Shenzhen}

  \icmlcorrespondingauthor{Qiming Li}{qmli@ir.hit.edu.cn}
  

  \icmlkeywords{Machine Learning, ICML}

  \vskip 0.3in
]
\printAffiliationsAndNotice{\icmlEqualContribution}



\begin{abstract}
{Although Large Vision-Language Models (LVLMs) have demonstrated remarkable performance on downstream tasks, they frequently produce contents that deviate from visual information, leading to object hallucination. To tackle this, recent works mostly depend on expensive manual annotations and training cost, or decoding strategies which significantly increase inference time. In this work, we observe that LVLMs' attention to visual information is significantly enhanced when answering caption queries compared to non-caption queries. Inspired by this phenomenon, we propose \textbf{C}aption-guided Visual \textbf{A}ttention \textbf{St}eering (\textbf{CAST}), a training-free, plug-and-play hallucination mitigation method that leverages the attention activation pattern corresponding to caption queries to enhance LVLMs' visual perception capability. Specifically, we use probing techniques to identify attention heads that are highly sensitive to caption queries and estimate optimized steering directions for their outputs. This steering strengthens LVLM's fine-grained visual perception capabilities, thereby effectively mitigating object hallucination. CAST reduced object hallucination by an average of 6.03\% across five widely used LVLMs and five benchmarks including both discriminative and generative tasks, demonstrating state-of-the-art performance while adding little inference cost and preserving other foundational capabilities.}
\end{abstract}

\section{Introduction}

Despite the remarkable performance of Large Vision-Language Models (LVLMs) on downstream tasks, it is widely observed that they frequently generate content that conflicts with the corresponding visual input, leading to object hallucination \citep{sahoo2024comprehensive,huang2023survey}.
\begin{figure}[t]
  \includegraphics[width=\columnwidth]{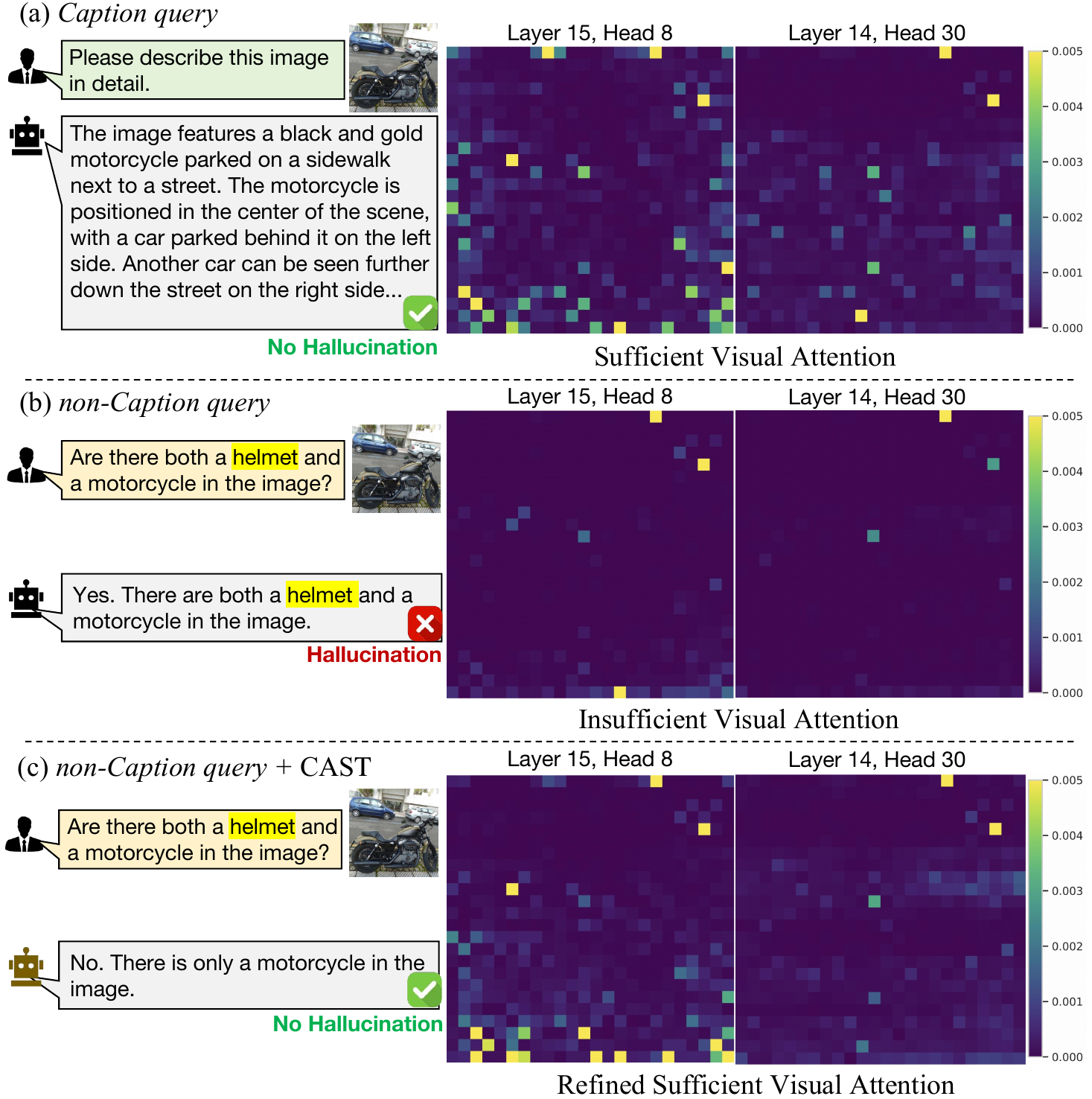}
  \caption{The visualization of attention weights at image patch level across different conversation settings. LLaVA-1.5-7b correctly generates the detailed content of the image in response to the caption query, but exhibits hallucination (e.g., "helmet") when answering the non-caption query. CAST refines LVLM's visual attention patterns from insufficient to sufficient, effectively enhancing visual perception capability and mitigating object hallucination.
  }
  \label{fig:figure1}
\end{figure}

To tackle this, recent works for mitigating hallucination mostly use contrastive decoding strategies \citep{leng2024mitigating,zhong2024investigatingmitigatingmultimodalhallucination} which arise high inference latencies, or training LVLMs using carefully designed data \citep{you2023ferret,yu2024rlhf} which incurs expensive manual annotation and computation cost. Furthermore, interpretability studies \citep{arif2025fixing,Bi2024UnveilingVP} have identified insufficient attention to visual information as an underlying cause of hallucination.
To address the aforementioned limitations and the underlying cause, we focus on exploring how to enhance LVLMs' perception capability by providing sufficient attention to visual information, without modifying model parameters or introducing significant inference cost.

In this work, we observe that caption query (e.g."Please describe this image in detail.") is a special type of instruction that plays a critical role in LVLM's pre-training stage for text-image alignment, endowing the model with fine-grained visual perception capability.
Furthermore, as shown in Figure \ref{fig:figure1} (a) and (b), we reveal a critical phenomenon: visual attention across particular attention heads is significantly enhanced when fed caption queries versus non-caption queries. We term these attention heads as \textit{caption-guided attention heads}. 
As an enhancement of their visual attention is accompanied by a reduction in object hallucination, it may indicate that these heads are responsible for the fine-grained perception capabilities.
Inspired by this phenomenon, we propose \textbf{C}aption-guided Visual \textbf{A}ttention \textbf{St}eering (\textbf{CAST}), a training-free, plug-and-play method, which probes and refines caption-guided attention heads outputs during inference to enhance LVLM‘s fine-grained visual perception capability and mitigate object hallucination.
Specifically, our method unfolds in three steps. First, following prior work \citep{li2024inference}, we use probing techniques to identify these caption-guided attention heads.
Furthermore, we compute attention output shift vectors for these attention heads, which quantify the output differences from non-caption to caption queries and serve as a fine-grained perception optimization direction.
Finally, we apply the precomputed shift vectors to steer caption-guided attention heads during inference, steering their outputs toward a state optimized for fine-grained visual perception and effectively mitigating object hallucinations. 
As shown in Figure \ref{fig:figure1} (b) and (c), \textbf{CAST} leads to a notable enhancement in visual attention and effectively mitigates object hallucination.

Consistent improvement across five widely used LVLMs and five benchmarks demonstrates that \textbf{CAST} achieves state-of-the-art (SOTA) performance. On the POPE \citep{li2023evaluating} benchmark, the accuracy and the F1 score improve by 5.14\% and 5.50\% on average. Furthermore, hallucination rates decrease by 7.8\% on the MMHalBench \citep{sun2023aligning}, while the informativeness of the responses improves.  

In summary, our main contributions are three-fold:
\begin{itemize}[leftmargin=*]
    \item Our work is the first to explicitly reveal the impact of caption queries versus non-caption queries on the attention activation patterns of LVLMs, providing novel insights for the optimization of visual attention.
    \item We propose \textbf{CAST}, a training-free method that effectively mitigates object hallucination by refining caption-guided attention head outputs during inference with little additional inference cost.
    \item Comprehensive experimental results demonstrate that \textbf{CAST} not only mitigates hallucination effectively but also shows strong generalization, preserving LVLM's other foundational capabilities.

\end{itemize}

\begin{figure*}[h]
  \centering
  \includegraphics[width=0.95\linewidth]{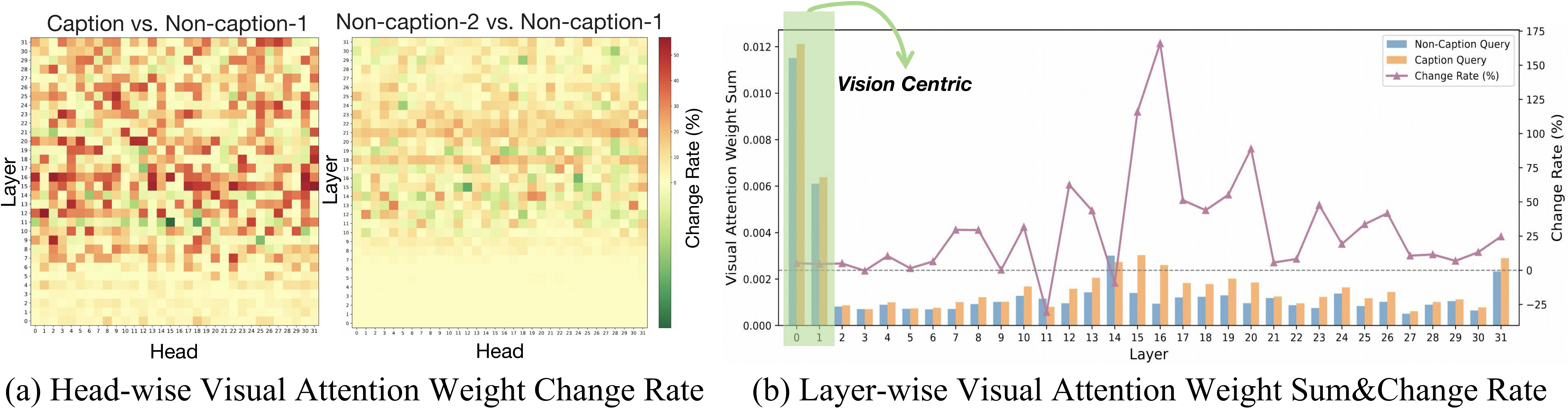}
  \caption{A quantitative analysis from head-wise (a) and layer-wise (b) perspective on visual attention weights, which demonstrates that caption queries significantly enhance visual attention of LLaVA-1.5-7b. 
  }
  \label{fig:quan}
\end{figure*}

\section{Related Works}

\subsection{Large Vision-Language Models}
Recent advances in large vision-language models have significantly pushed the boundaries of multimodal understanding and generation.
Several powerful LVLMs based on open-source LLM backbones combined with visual encoders have achieved impressive capabilities through vision-language pretraining. Furthermore, recent searches have further improved model performance by employing high-resolution visual encoders \citep{hong2024cogvlm2} and exploring reinforcement learning methods, such as RLHF \citep{yu2024rlhf}. Closed-source models, such as GPT-4o \citep{hurst2024gpt} and Gemini 1.5 \citep{reid2024gemini} have demonstrated even more powerful performance. In addition, a growing body of work emphasizes scaling strategies, cross-modal alignment, and integration of external knowledge sources, which further enrich the reasoning and generation abilities of LVLMs. However, despite these advances, recent LVLMs still suffer from hallucination problems, and addressing how to cost-effectively mitigate hallucination remains an important open question that demands deeper exploration.

\subsection{Mitigating Hallucination in LVLMs}
Current methods for mitigating hallucination in LVLMs can be broadly categorized into two types: data-driven training methods and training-free methods. Training-based methods typically involve introducing novel training objectives \citep{chen2024alleviating} and utilizing carefully curated datasets \citep{gunjal2024detecting,liu2023mitigating,yu2024rlaif,you2023ferret}. For training-free methods, the main strategies include designing decoding techniques \citep{leng2024mitigating,chen2024halc,chuang2023dola,huang2024opera,zhong2024investigatingmitigatingmultimodalhallucination} during the inference phase and leveraging language or visual prompts \citep{lee2023volcano,an2024agla}. 
PAI \citep{liu2024paying} intervenes in attention heads by leveraging the direction and magnitude of their original outputs, and optimizes the output distribution during decoding to mitigate hallucinations. VTI \citep{liu2024reducing} reduces hallucinations by steering hidden states during inference to enhance the stability of vision features. Beyond these approaches, a number of studies highlight the importance of understanding the underlying mechanisms that trigger hallucinations, suggesting that architectural and interpretability-driven steering may offer complementary solutions. However, our work is the first to explicitly reveal the impact of caption queries on the attention activation patterns of LVLMs and mitigate hallucination by applying caption-guided attention head steering during inference.

\section{Analysis of Caption Queries' Effect on Visual Attention}

We performed a quantitative analysis to validate the primary motivation for CAST: caption queries uniquely refine visual attention patterns in LVLMs in a way that other queries do not. Using a sample of 1,000 images from the MS-COCO dataset \citep{lin2014microsoft}, we designed three distinct queries for each image to analyze the effect of query type: one caption query and two vision-oriented non-caption queries with distinct meanings (non-caption-1 \& non-caption-2). To quantify the effect on visual attention for caption and non-caption queries, we compute the $\mathbf{Change\ Rate}$ of attention weights across all layers and attention heads. Further details on this computation and the experimental setup are available in Appendix \ref{sec:B}.

Experimental results in Figure \ref{fig:quan} indicate that caption queries demonstrate significant enhancements on LVLM's visual attention weights compared with non-caption queries, especially in the mid layers. As shown in Figure \ref{fig:quan} (a), 65.92\% of attention heads, which are concentrated primarily in middle layers, exhibit increased visual attention weights when fed caption queries. As shown in Figure \ref{fig:quan} (b), 30 out of 32 layers exhibit a consistent enhancement in visual attention. 
Notably, the mid-layer attention heads demonstrate the most substantial improvements, which indicates their critical role in enabling LVLMs’ fine-grained perception capability.
Our analysis provides clear feasibility and insights for locating and refining attention heads by leveraging the visual attention enhancement induced by caption queries to mitigate object hallucinations.

\section{Methods}


\subsection{Preliminaries: The Transformer Residual Stream}
 
We consider a LVLM parametrized by \( \theta \). The model receives as input a visual input $ \boldsymbol{V} = \{v_1, v_2, \dots, v_m\} $ and a textual query $ \boldsymbol{T} = \{t_1, t_2, \dots, t_n\} $, where $ m $ and $ n $ denote the sequence lengths of the visual input and textual inputs. The textual and visual inputs are concatenated together to form the first layer input $ \boldsymbol{H}^{1} = \mathrm{concat}(\boldsymbol{V},\boldsymbol{T}) \in \mathbb{R}^{(m+n)\times d}$ for the $L$ layers $\times$ $H$ heads language decoder. 

\begin{figure*}[!ht]
  \centering
  \includegraphics[width=1\linewidth]{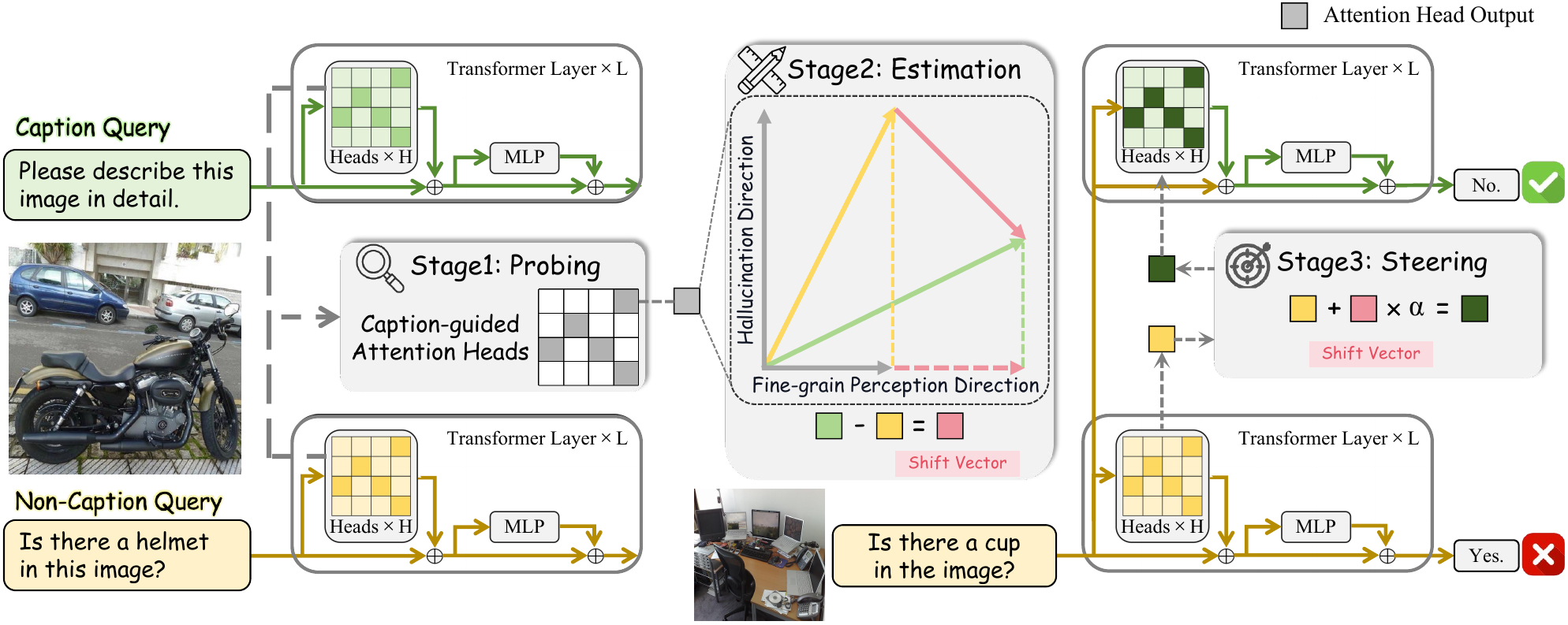}
  \caption{An overview of the CAST method. Each square in the matrix represents the attention head output. Squares with dark green color indicate refined caption-guided attention head outputs. CAST consists of three stages: \textbf{(1) Caption-Guided Attention Heads Probe \S\ref{probe}:} We use probing techniques to identify caption-guided attention heads, which exhibit enhanced visual attention when fed caption queries versus non-caption queries. \textbf{(2) Estimation of Perception Refined Vectors \S\ref{vector}:} We estimate the perception refined vectors by computing the attention output shift vectors from feeding non-caption queries to caption queries. \textbf{(3) Steering at Inference Time \S\ref{intervention}:} We apply the precomputed attention refined vectors to the Top-$K$ caption-guided attention heads during inference, thereby enhancing visual attention and activating the model's inherent fine-grained visual perception capability and effectively mitigate object hallucination. }
  \label{fig:method}
\end{figure*}

During the forward pass, the input $ \boldsymbol{H}^{l}$ received by the $h$-th attention head at $l$-th layer is linearly transformed using independent weight matrices to generate the Query, Key and Value matrices, denoted as $\boldsymbol{Q}_{(l,h)}\in\mathbb{R}^{(m+n)\times d}$, $\boldsymbol{K}_{(l,h)}\in\mathbb{R}^{(m+n)\times d}$ and
$\boldsymbol{V}_{(l,h)}\in\mathbb{R}^{(m+n)\times d}$,
where $d$ denotes the head-specific hidden dimension. The generated Query, Key, and Value matrices are then used to compute the attention score, attention weight matrix, and attention output as follows: 
\begin{equation}
\boldsymbol{\dot{A}}_{(l,h)} = \frac{\boldsymbol{Q}_{(l,h)} \boldsymbol{K}_{(l,h)}^T}{\sqrt{d}},
\boldsymbol{A}_{(l,h)} = \textrm{softmax}(\boldsymbol{\dot{A}}_{(l,h)}+\boldsymbol{M}),
\label{eq:attention_relevance}
\end{equation}

\begin{equation}
    \boldsymbol{M}[i,j]=
    \begin{cases}
0 & \text{if } j \leq i \\
-\infty & \text{if } j > i
\end{cases}
\end{equation}

\begin{equation}
\boldsymbol{O}_{(l,h)} = \boldsymbol{A}_{(l,h)}\boldsymbol{V}_{(l,h)},
\label{eq:attention_output}
\end{equation}
where $\boldsymbol{M}$ is the causal mask matrix. At each layer, the hidden states pass through multi-head attention (MHA), which comprise $H$ independent attention heads, each performing separate linear transformations. Specifically, the MHA mechanism can be formulated as: 
\begin{equation}
    \boldsymbol{H}^{l+1} = \boldsymbol{H}^{l} + \sum_{h=1}^{H} \boldsymbol{O}_{(l,h)} \cdot \boldsymbol{W}_o^l,
\end{equation}
where $\boldsymbol{W}_o^l\in\mathbb{R}^{Hd\times d}$ is the learnable weight matrix and maps d-dimensional attention outputs of heads into hidden state representations, which are then fed into a standard multilayer perception (MLP) for further processing.
Finally, the model predicts the next token in auto-regressive manner. 

\subsection{Caption-Guided Attention Heads Probe}
\label{probe}

This module aims to identify caption-guided attention heads, which are also visually sensitive and exhibit significant differences in attention outputs when responding to caption and non-caption queries.
Since LVLMs generate tokens in an auto-regressive manner, CAST focuses on the attention matrices of the last input token, $\boldsymbol{\dot{A}}_{(l,h)}[m+n]$, which aggregates the most comprehensive visual and textual information.
Furthermore, we aim to capture the differences in attention activation patterns when fed caption queries versus non-caption queries, as well as minimize the influence of textual semantic information during probing.
To achieve this, we mask $\boldsymbol{\dot{A}}_{(l,h)}[m+n]$ to exclude attention towards all textual tokens during the forward pass, and compute the modified attention output:
\begin{equation}
    \boldsymbol{\hat{M}}[i,j]=
    \begin{cases}
0 & \text{if } j \leq i \\
-\infty & \text{if } j > i \  or \ (i=m+n \ and \ j>n)
\end{cases}
\end{equation}
\begin{equation}
   {\boldsymbol{\hat{O}}_{(l,h)} }=  \textrm{softmax}({\boldsymbol{\dot{A}}_{(l,h)}+\boldsymbol{\hat{M}}})\boldsymbol{V}_{(l,h)},
\end{equation}

\begin{equation}
    {\boldsymbol{\widetilde{O}}_{(l,h)} =  {\boldsymbol{\hat{O}}_{(l,h)} }[m+n]}.
\end{equation}

For a dataset with a batchsize of $B$, the last token's modified attention output of $b$-th VQA problem when answering caption query and non-caption query are denoted as $\boldsymbol{\widetilde{O}}^{b}_{(l,h)}$ and $\boldsymbol{\widetilde{O'}}^{b}_{(l,h)}$.
For each attention head $Head_{(l,h)}$, we use $B$ pairs of modified attention output
to train a binary classifier $f_{l,h}(\cdot)$ that predicts whether the input sentence is a caption query. Finally, we select the attention heads with the Top-$K$ highest classification accuracy as the caption-guided attention heads. The formulas are summarized as:
\begin{equation}
f_{l,h}^*=\mathop{\arg\min}\limits_{{f_{l,h(\cdot)}}}\sum_{b=1}^{B} \mathcal{L}\left(f_{l,h}\left(x_b\right), y_b\right),
\end{equation}
\begin{equation}
    Heads = \{ Head_{(l,h)} \mid Head_{(l,h)} \in \operatorname{TopK}(\operatorname{Acc}(f_{l,h}^*)) \}
\end{equation}
where $f_{l,h}^*$ denotes the final probe, $\mathcal{L}$ denotes the loss function of the probes, $x_b\in\{\boldsymbol{\widetilde{O}}^{b}_{(l,h)} ,\boldsymbol{\widetilde{O'}}^{b}_{(l,h)}\}$ denotes the input of the classifier, $y_b\in\{0,1\}$ denotes the category of query (0 for caption query, 1 for non-caption query, respectively), and $K$ denotes the number of selected heads.

\subsection{Estimation of Perception Refined Vectors}
\label{vector}
This module aims to use caption-guided attention heads to accurately estimate the perception refined vectors.
For a dataset with a batchsize of $B$, the last token's origin attention output of $b-$th VQA problem when answering caption query and non-caption query are denoted as $\boldsymbol{{O}}^{b}_{(l,h)}$ and $\boldsymbol{{O'}}^{b}_{(l,h)}$. 
To estimate the fine-grained perception direction for each attention head, attention output shift vector is computed as follows:
\begin{equation}
   \boldsymbol{S}_{(l,h)} = \frac{1}{B} \sum_{b=1}^{B}\left( \boldsymbol{O}_{(l,h)}^b - \boldsymbol{O'}^{b}_{(l,h)}\right).
\end{equation}
These shift vectors estimate the visual attention difference between caption queries and non-caption queries, which serve as the fine-grained perception directions. In particular, the modified attention outputs $\boldsymbol{\widetilde{O}}^{b}_{(l,h)}$, $\boldsymbol{\widetilde{O'}}^{b}_{(l,h)}$ are not used to estimate the refined vectors, as these values are not directly derived from the original inference process. In contrast, using the original attention outputs leads to more robust refined vectors.

\subsection{Steering at Inference Time}
\label{intervention}
This module aims to refine caption-guided attention heads at inference time. We leverage the precomputed refined vectors to steer these heads from insufficient visual attention states to sufficient states, thereby enhancing the model’s fine-grained visual perception capability and mitigate hallucination. At each layer, the updated hidden state after steering is computed as:
\begin{equation}
    \boldsymbol{H}^{l+1} = \boldsymbol{H}^{l} + \sum_{h=1}^{H} \left(\boldsymbol O_{(l,h)}+\mathbb{I}_{(l,h)}\alpha\boldsymbol{S}_{(l,h)}\right) \cdot \boldsymbol{W}_o^l,
\end{equation}
where $\mathbb{I}_{(l,h)}$ is a gating function, assigning a value of $1$ to caption-guided attention heads, and 0 to the others. $\alpha$ represents the intensity of the steering.

In conclusion, CAST significantly enhances LVLM's fine-grained perception capability, which is attributed to the unique role of caption queries during the pre-training stage for text-image alignment, and their sufficient visual attention patterns. Furthermore, CAST benefits from the inference-time steering paradigm, which provides an inherent advantage in inference latency. 

\section{Experiments}
\subsection{Experimental Setup}
\label{Setup}
\noindent\textbf{Benchmarks.}
We evaluate our proposed CAST method across five benchmarks, including discriminative and generative tasks to measure its effectiveness and robustness. See Appendix \ref{benchmarks} for details of benchmarks and Appendix \ref{more_bench}, \ref{sec:F} for more advanced and domain-specific benchmarks.

\begin{table*}[!ht]
\small
\renewcommand{\arraystretch}{1}
\centering
\setlength{\tabcolsep}{4.5pt}
\caption{Main results on POPE tasks. The best performances are bolded.}
\begin{tabular}{clcccccc} 
\toprule
\multirow{2}{*}{\textbf{Setting}} & \multirow{2}{*}{\textbf{Method}} & \multicolumn{2}{c}{\textbf{LLaVA-1.5-7b}} & \multicolumn{2}{c}{\textbf{Qwen-VL-Chat}} & \multicolumn{2}{c}{\textbf{LLaVA-NeXT}}\\
\cmidrule(lr){3-4} \cmidrule(l){5-6} \cmidrule(l){7-8}
& & Accuracy$\uparrow$ & F1-Score$\uparrow$ & Accuracy$\uparrow$ & F1-Score$\uparrow$ &Accuracy$\uparrow$ & F1-Score$\uparrow$\\
\midrule
\multirow{6}{*}{Random} 
&\cellcolor{gray!15} Regular & \cellcolor{gray!15}83.29 & \cellcolor{gray!15}81.33 & \cellcolor{gray!15}84.63 & \cellcolor{gray!15}82.61 & \cellcolor{gray!15}84.78 & \cellcolor{gray!15}86.43 \\
& VCD & 87.73 & 87.16 & 86.93 & 85.46 & 88.76 & 89.57 \\
& OPERA & 89.20 & 88.81 & 85.71 & 84.64 & 90.27 & 89.71 \\
& PAI & 86.33 & 84.56 & 85.38 & 85.54 & 88.40 & 87.16 \\
& VTI & 89.50 & 88.89 & 86.73 & 85.59 & 89.23 & 88.68 \\
& \textbf{CAST(ours)} & \textbf{89.87} \textcolor{darkgreen}{(+6.58)} & \textbf{89.43} \textcolor{darkgreen}{(+8.10)} & \textbf{88.17} \textcolor{darkgreen}{(+3.54)} & \textbf{87.31} \textcolor{darkgreen}{(+4.70)} & \textbf{90.68} \textcolor{darkgreen}{(+5.90)} & \textbf{90.42} \textcolor{darkgreen}{(+3.99)}\\
\cmidrule(lr){2-8}
\multirow{6}{*}{Popular}
&\cellcolor{gray!15}Regular &\cellcolor{gray!15} 81.88 &\cellcolor{gray!15} 80.06 &\cellcolor{gray!15} 83.63 &\cellcolor{gray!15} 81.53 & \cellcolor{gray!15}83.23 &\cellcolor{gray!15} 84.77 \\
& VCD & 85.38 & 85.06 & 85.17 & 83.68 & 87.01 & 87.70 \\
& OPERA & 86.64 & 86.62 & 84.82 & 83.99 & 87.16 & 87.68 \\
& PAI & 85.33 & 83.62 & 84.20 & 83.10 & 86.65 & 86.99 \\
& VTI & 87.36 & 86.69 & 85.67 & 84.48 & 87.33 & 87.16 \\
&  \textbf{CAST(ours)} & \textbf{88.32} \textcolor{darkgreen}{(+6.44)} & \textbf{87.95} \textcolor{darkgreen}{(+7.89)} & \textbf{87.73} \textcolor{darkgreen}{(+4.10)} & \textbf{86.84} \textcolor{darkgreen}{(+5.31)} & \textbf{89.53} \textcolor{darkgreen}{(+6.30)} & \textbf{89.24} \textcolor{darkgreen}{(+4.47)} \\
\cmidrule(lr){2-8}
\multirow{6}{*}{Adversarial}
&\cellcolor{gray!15}Regular &\cellcolor{gray!15} 78.96 &\cellcolor{gray!15} 77.57 &\cellcolor{gray!15} 81.03 &\cellcolor{gray!15} 79.30 &\cellcolor{gray!15} 81.19 &\cellcolor{gray!15} 82.50 \\
& VCD & 80.88 & 81.33 & 83.10 & 82.04 & 84.80 & 85.23 \\
& OPERA & 81.24 & 81.38 & 82.67 & 79.89 & 85.20 & 85.54 \\
& PAI & 83.17 & 81.67 & 82.19 & 82.06 & 84.32 & 83.68 \\
& VTI & 82.57 & 82.11 & 83.13 & 82.16 & 85.35 & 84.52 \\
& \textbf{CAST(ours)} & \textbf{84.27} \textcolor{darkgreen}{(+5.31)} & \textbf{84.41} \textcolor{darkgreen}{(+6.84)} & \textbf{84.33} \textcolor{darkgreen}{(+3.30)} & \textbf{83.92} \textcolor{darkgreen}{(+4.62)} & \textbf{85.97} \textcolor{darkgreen}{(+4.78)} & \textbf{86.07} \textcolor{darkgreen}{(+3.57)} \\
\bottomrule
\end{tabular}
\label{tab:main_result}
\end{table*}

\noindent\textbf{Baselines.} We adopt LLaVA-1.5-7b, Qwen-VL-Chat, LLaVA-NeXT \citep{liu2024llavanext} as baseline LVLMs, compared with several SOTA training-free methods. See Appendix \ref{ad_models} for results on more advanced LVLMs, Appendix \ref{sec:more_methods} for results compared with other SOTA methods.

\textbf{(1) Baselines tailored for decoding:} 
VCD \citep{leng2024mitigating} contrasts model logits derived from original and distorted visual input to reduce the over-reliance on statistical bias. OPERA \citep{huang2024opera} introduces a penalty term on the logits during the beam-search decoding to mitigate the over-trust issue.

\textbf{(2) Baselines utilizing inference-time intervention (ITI):} 
PAI \citep{liu2024paying} intervenes on attention heads by leveraging their original direction and optimizes the output distribution during decoding to mitigate language bias. VTI \citep{liu2024reducing} mitigates hallucination by steering hidden states at inference to enhance the stability of visual features. 

Despite prior findings \citep{bi2024unveiling} indicating the significant role of attention heads in visual perception, there is a lack of approaches that analyze at head level and do not rely on specific decoding strategies (which increase inference time). The idea of using the attention differential between caption and non-caption inputs to guide inference interventions distinguishes CAST from earlier ITI works.

\noindent\textbf{Implementation Details.} In our experiments, we utilize 1000 task-diverse VQAs from LLaVA-1.5-7b pretraining dataset, each paired with a specific caption query, to identify caption-guided attention heads and compute the attention shift vectors. For each attention head, an SVM \citep{cortes1995support} is employed as the classifier, with further discussion provided in Appendix \ref{sec:impacts}. CAST incorporates steering across all model layers; refer to Appendix \ref{layers} for further details and Appendix \ref{detailed_ablation} for hyperparameter configurations. 
More details of the experimental setup are provided in Appendix \ref{sec:C}.

\subsection{Main Results}
\begin{figure*}[!ht]
    \centering
    \includegraphics[width=1\linewidth]{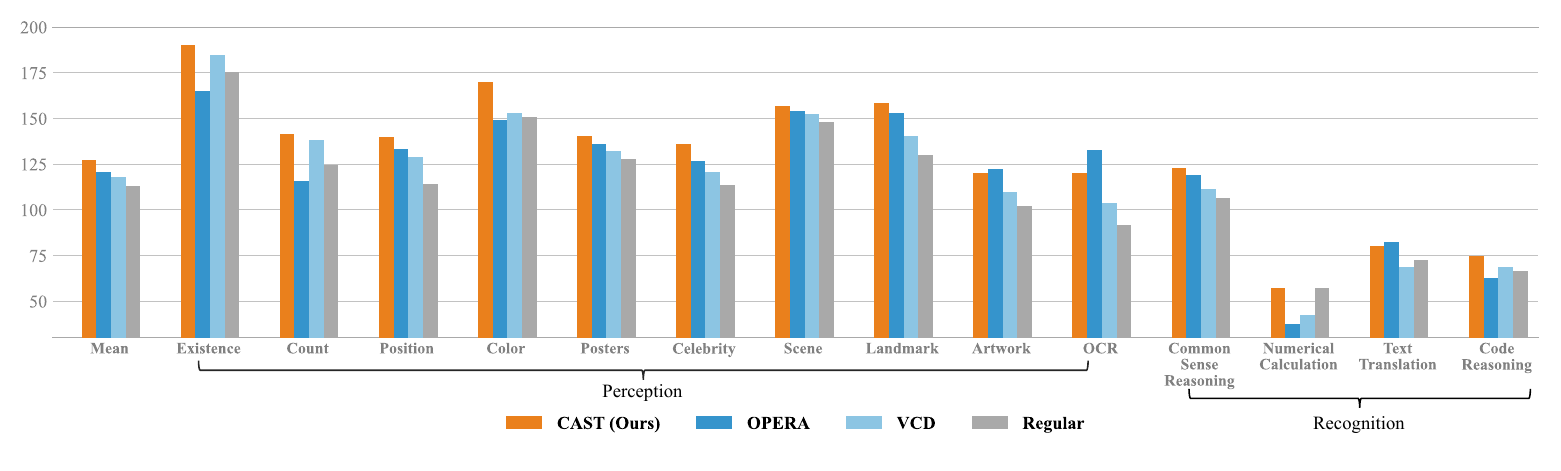}
    \caption{Main results of LLaVA-1.5-7b on the MME.}
    \label{fig:enter-label}
\end{figure*}

As shown in Table \ref{tab:main_result}-\ref{tab:mmhbench-result} and Figure \ref{fig:enter-label}, we summarize our main findings as follows:

\begin{table*}[!ht]
\centering
\small
\renewcommand{\arraystretch}{1}

\begin{minipage}{0.48\textwidth}
\centering
\setlength{\tabcolsep}{0.3pt} 
\caption{Results on CHAIR benchmark. Max new tokens are set to be 512.}
\begin{tabular}{lcccccccc}
\toprule
\multirow{2}{*}{\textbf{Method}} & \multicolumn{4}{c}{\textbf{LLaVA-1.5-7b}} & \multicolumn{4}{c}{\textbf{Qwen-VL-Chat}} \\
\cmidrule(lr){2-5}
\cmidrule(lr){6-9}
& $C_S$ \textbf{$\downarrow$} & $C_I$ \textbf{$\downarrow$} & Recall$\uparrow$& Len& $C_S$ \textbf{$\downarrow$} &  $C_I$ \textbf{$\downarrow$} &Recall$\uparrow$ & Len \\
\midrule
\rowcolor{gray!15}Regular & 52.8 & 15.9 &77.3 &93.4& 2.8 & 3.0&31.0&5.3 \\
VCD & 51.0 & 14.9 &77.2& 101.9& 1.4 & 1.2&30.8&4.0  \\
OPERA & 45.6 & 13.1& \textbf{78.5}&95.3 & 1.7 & 1.3&31.9 &4.4\\
PAI & 38.3 & 12.4 & 76.9 &94.4 &  1.3 & 1.2&\underline{32.2}&4.2 \\
VTI & \underline{36.9} & \underline{12.1} &76.8&93.8 &\underline{1.1} &\underline{1.1}&31.4 &4.2\\
\textbf{CAST} & \textbf{34.6} & \textbf{11.5} &\underline{78.2}&95.8&  \textbf{1.0} & \textbf{0.9} &\textbf{32.6} &4.4 \\
\bottomrule
\end{tabular}

\label{tab:chair-result}
\end{minipage}
\hfill
\begin{minipage}{0.48\textwidth}
\centering
\setlength{\tabcolsep}{1.2pt} 
\caption{Results on MMHal-Bench and MHumanEval (evaluated by GPT-4 \& Human).}
\begin{tabular}{lcccccc}
\toprule
\multirow{2}{*}{\textbf{Method}} & \multicolumn{3}{c}{\textbf{LLaVA-1.5-7b}} & \multicolumn{3}{c}{\textbf{Qwen-VL-Chat}} \\
\cmidrule(lr){2-4}
\cmidrule(lr){5-7}
& Score\textbf{$\uparrow$} & VH.\%\textbf{$\downarrow$} & Hu.\%\textbf{$\downarrow$} 
& Score\textbf{$\uparrow$} & VH.\%\textbf{$\downarrow$} & Hu.\%\textbf{$\downarrow$} \\
\midrule
\rowcolor{gray!15}Regular & 1.86 & 63.5 & 67.1 & 2.93 & 41.1 & 61.0 \\
VCD & 2.12 & 54.2 & 66.7 & 2.77 & 39.2 & 61.5 \\
OPERA & 2.15 & 54.2 & 63.0 & 2.94 & 38.4 & 58.2 \\
PAI & 2.27 & 53.2 & \underline{62.5} & 2.87 & 39.5 & \underline{56.7} \\
VTI & \underline{2.33} & \underline{52.2} & 63.4 & \underline{2.99} & \underline{38.4} & 57.4 \\
\textbf{CAST} & \textbf{2.43} & \textbf{51.0} & \textbf{61.5} 
& \textbf{3.04} & \textbf{38.0} & \textbf{56.0} \\
\bottomrule
\end{tabular}

\label{tab:mmhbench-result}
\end{minipage}
\end{table*}

\begin{table*}[!ht]
\centering
\renewcommand{\arraystretch}{0.1}
\setlength{\tabcolsep}{5pt}
\caption{We construct a caption query candidate pool (N=16), where we derive our test cases as follows: (1) four queries are randomly selected; (2) one optimal query is chosen using caption query optimization algorithm; and (3) an ensemble steering strategy is applied. \textbf{VA} (\%) indicates the average percentage of attention weights over visual tokens when fed corresponding query. $\alpha$ and $K$ denote the intensity and number of the steering. We select the optimal parameters separately for each setting. }
\begin{tabular}{l|c|cc|cc|cc|cc|cc}
\toprule
\multirow{2}{*}{\textbf{Setting}} 
& \multirow{2}{*}{\textbf{VA (\%)}} 
& \multicolumn{2}{c|}{\textbf{Parameters}} 
& \multicolumn{2}{c}{\textbf{Random}} 
& \multicolumn{2}{c}{\textbf{Popular}} 
& \multicolumn{2}{c}{\textbf{Adversarial}} 
& \multicolumn{2}{|c}{\textbf{Average}} \\
\cmidrule(lr){3-4} \cmidrule(lr){5-6} \cmidrule(lr){7-8} \cmidrule(lr){9-10} \cmidrule(lr){11-12}
& & \textbf{$\boldsymbol\alpha$} & \textbf{$\boldsymbol{K}$} & \textbf{ACC}$\uparrow$ & \textbf{F1}$\uparrow$ & \textbf{ACC}$\uparrow$ & \textbf{F1}$\uparrow$ & \textbf{ACC}$\uparrow$ & \textbf{F1}$\uparrow$ & \textbf{ACC}$\uparrow$ & \textbf{F1}$\uparrow$ \\
\midrule
\rowcolor{gray!20}
Regular     & 31.4 & -    & -    & 83.29 & 81.33 & 81.88 & 80.06 & 78.96 & 77.57 & 81.38 & 79.65 \\
\midrule
Random1     & 46.8 (+15.4) & 1.25  & 100    & 88.59 & 88.15 & 86.95 & 86.55 & 83.08 & 83.25 & 86.21 & 85.98 \\
Random2     & 45.6 (+14.2) & 1.50  & 100    & 88.65 & 88.21 & 87.01 & 86.68 & 83.15 & 83.33 & 86.27 & 86.07 \\
Random3     & 44.7 (+13.3) & 1.50  & 125    & 89.02 & 88.65 & 87.41 & 87.05 & 83.58 & 83.72 & 86.67 & 86.47 \\
Random4     & 44.2 (+12.8) & 1.50  & 100    & 89.15 & 88.82 & 87.53 & 87.21 & 83.66 & 83.80 & 86.78 & 86.61 \\
\midrule
\rowcolor{green!20}
Optimized of $N$        & 43.4 (+12.0) & 1.50  & 100    & \textbf{89.87} & \textbf{89.43} & \textbf{88.32} & \textbf{87.92} & \textbf{84.27} & \textbf{84.41} & \textbf{87.49} & \textbf{87.26} \\
\midrule
\rowcolor{yellow!20}
Ensemble of $N$   & -    & 1.50    & 100    & 88.93 & 88.68 & 87.46 & 86.91 & 83.78 & 83.56 & 86.72 & 86.38 \\
\bottomrule
\end{tabular}

\label{tab:query_comparison}
\end{table*}

\textbf{(1) SOTA hallucination mitigation performance\ } Our proposed CAST method achieves SOTA hallucination mitigation performance across both discriminative and generative tasks. On the POPE benchmark, CAST improves accuracy by an average of +5.64\% and F1 Score by +5.50\%. On the CHAIR benchmark, CAST reduces the average hallucination metrics ($C_S$ and $C_I$) by 6.43 points. On MMHal-Bench, CAST improves the average Score by +0.16, while reduces the average VH Rate by 2.95\% and the Hu. Rate by 2.25\%. As shown in Appendix \ref{evidence}, CAST substantially mitigates the “yes-bias”, providing deeper evidence of CAST’s effectiveness in discriminative settings.

\textbf{(2) Generalizability across architectures and datasets\ }
CAST exhibits strong generalization capability across both model architectures and data sources. From the architectural perspective, CAST remains effective across models with different attention mechanisms, including those with optimized implementations such as Qwen-VL-Chat. This is because CAST stems from the difference in attention patterns between caption and non-caption queries, rather than the specific implementation details of the multi-head attention mechanism. From the data perspective, although the probing and refined vectors are computed using 1,000 samples from the LLaVA-1.5-7b pre-training dataset, they generalize well to other out-of-domain benchmarks and advanced LVLMs. These results highlight the generalizability across model architectures and datasets.

\textbf{(3) Preservation of foundational capabilities\ }
CAST not only mitigates hallucination but also preserves the LVLM’s other foundational capabilities. On the MME benchmark, CAST improves performance on all tasks, preserving most of LVLM’s foundational capabilities. Furthermore, CAST improves the informativeness score by 0.16 on MMHal-Bench, demonstrating that CAST effectively mitigates object hallucination without compromising informativeness.

\begin{figure*}[t]
    \centering
    \includegraphics[width=0.48\textwidth]{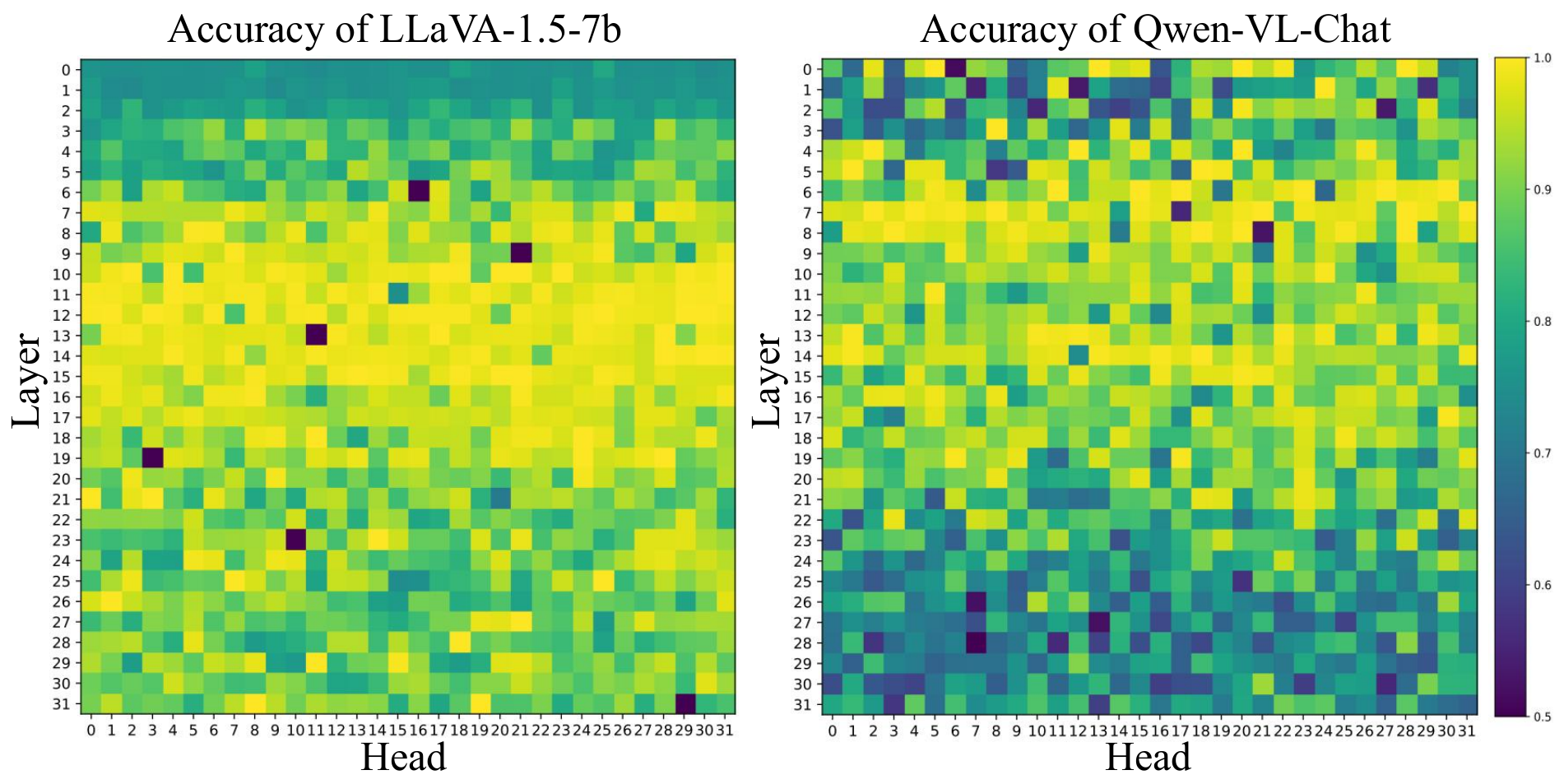}\hfill
    \includegraphics[width=0.48\textwidth]{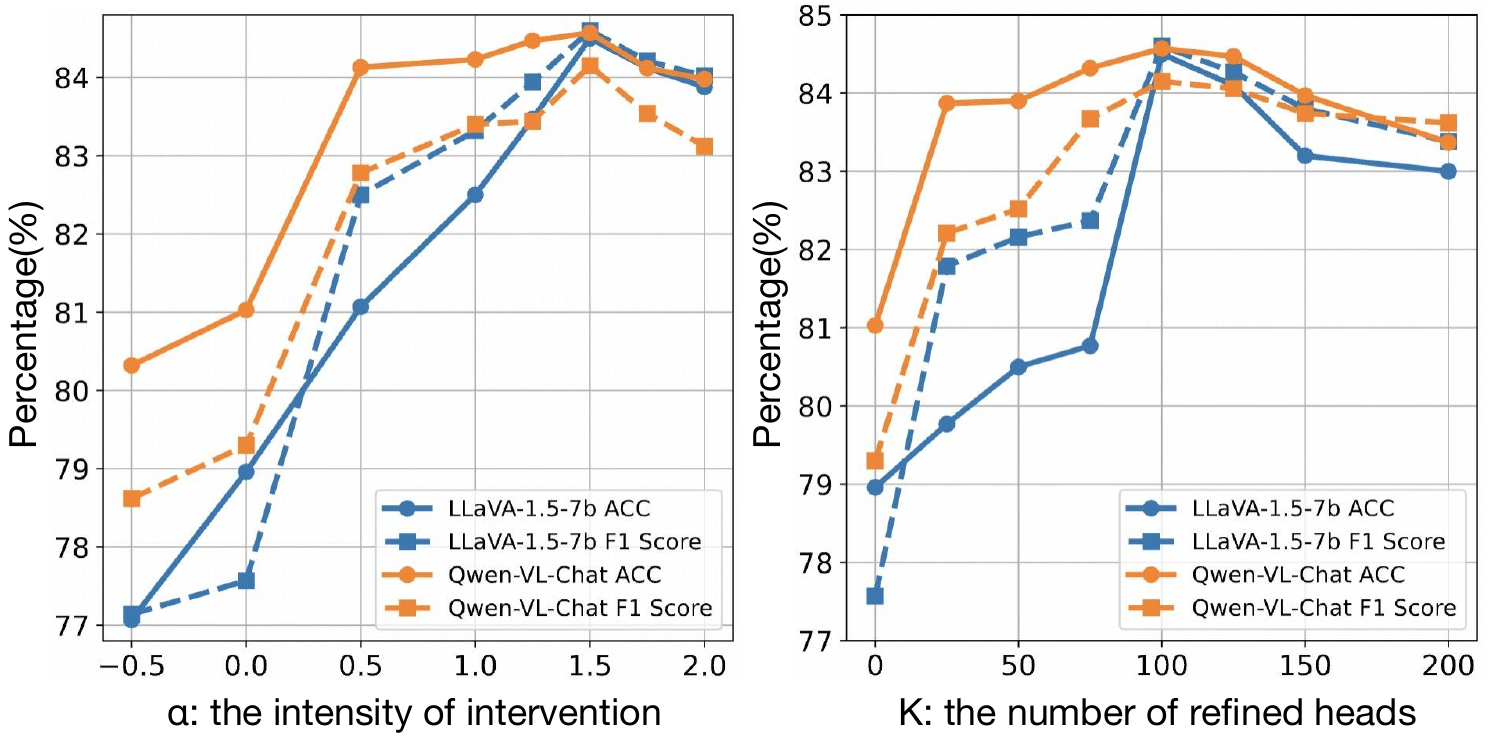}
    \caption{The accuracies of probes (left) and ablation study of $\alpha$ and $K$ on POPE (right).}
    \label{fig:double}
\end{figure*}

\section{Analysis and Discussions}
\subsection{Optimization via Caption Queries' Diversity}

To further enhance the robustness of CAST, we aim to leverage the diversity of caption queries and introduce two optimization strategies to improve real-world application.

\textbf{Candidate Caption Query Pool Expansion:} Caption queries refer to prompts with explicit semantics (e.g., "Please describe this image in detail") and strong cross-model transferability, which can be easily sourced from open pre-training datasets or generated using large language models. By expanding the candidate pool, we increase the diversity and generalizability of caption-guided attention heads probing.

\textbf{Caption Query Optimization Algorithm:} Our experiments reveal that the shift cost—the attention weights change from a non-caption query to a caption query on a dataset—varies when fed different caption queries. Caption queries with minimal necessary shift cost yield better hallucination mitigation performance and we term these queries as optimized queries. This is possibly because optimized queries require less attention diversion from textual to visual information while still enabling fine-grained perception capability. As a result, optimized queries preserve the model's native attention distribution better and strike a balance between visual and textual attention. As shown in Table \ref{tab:query_comparison}, by expanding the pool of candidate caption queries and applying the proposed caption query optimization algorithm, we can further enhance CAST's performance.

\textbf{Multi-query Feature Ensemble Algorithm:}
Although CAST achieves stable performance across different caption queries, we propose a multi-query ensemble strategy to reduce the influence of suboptimal or outlier queries. Specifically, we integrate attention features from multiple caption queries to identify consistent caption-guided heads and estimate perception refined vectors. Strengthening these heads improves object hallucination mitigating performance and provides robust steering against individual prompt variability. As shown in Table \ref{tab:query_comparison}, while this ensemble may be marginally less optimal than using the optimized caption query, it substantially improves the reliability of CAST under various conditions.



\begin{table*}[!ht]
  \centering
  \begin{minipage}[t]{0.48\textwidth}
    \centering
    \small
    \renewcommand{\arraystretch}{1.38} 
    \setlength{\tabcolsep}{3pt} 
    \caption{Impact of over-steering on CHAIR benchmark. $Max\ new\ tokens$ is set to 64.}
    \begin{tabular}{l|cc|ccc}
      \toprule
      \textbf{Method} & \textbf{$C_S\downarrow$} & \textbf{$C_I\downarrow$} & \textbf{PPL} & \textbf{Coher. $\uparrow$} & \textbf{Fluency $\uparrow$}\\
      \midrule
      LLaVA-1.5-7b & 20.80 & 6.77 & 3.97 & 0.998 & 0.805 \\
      \midrule
      \quad + CAST & 17.20 & 5.50 & 4.11 & 0.998 & 0.791 \\
      \quad + CAST (over) & 18.60 & 6.00 & 4.23 & 0.997 & 0.809 \\
      \bottomrule
    \end{tabular}
    \label{tab:chair}
    \vspace{0pt} 
  \end{minipage}
  \hfill
  \begin{minipage}[t]{0.48\textwidth}
    \centering
    \small
    \renewcommand{\arraystretch}{1} 
    \setlength{\tabcolsep}{6pt} 
    \caption{Inference latency (TTFT, TPOT) and accuracy on POPE adversarial.}
    \begin{tabular}{lrrc}
      \toprule
      \textbf{Method} & \textbf{TTFT(ms)} & \textbf{TPOT(ms)} & \textbf{Acc(\%)}\\
      \midrule
      \rowcolor{gray!20} LLaVA-1.5-7b & 99.8 {1.0$\times$} & 36.0 {1.0$\times$} & 78.96 \\
      \midrule
      \quad + VCD & 160.1 {1.6$\times$} & 96.8 {2.7$\times$} & 80.88 \\
      \quad + PAI & 156.3 {1.6$\times$} & 93.6 {2.6$\times$} & 83.17 \\
      \midrule
      \rowcolor{green!20} \quad + CAST(ours) & 102.2 {1.0$\times$} & 36.5 {1.0$\times$} & 84.50 \\
      \bottomrule
    \end{tabular}
    \label{latency}
    \vspace{0pt}
  \end{minipage}
\end{table*}

\begin{table}[h]
    \small
    \setlength{\tabcolsep}{1pt} 
    \caption{Latency and performance comparisons between \textit{precomputed} and \textit{on-the-fly} approaches with CAST.}
    \begin{tabular}{lcccccc}
        \toprule
        \multirow{2}{*}{\textbf{Method}} & \multirow{2}{*}{\textbf{Latency}} & \multicolumn{3}{c}{\textbf{POPE} ($\uparrow$)} & \multicolumn{2}{c}{\textbf{CHAIR} ($\downarrow$)} \\
        \cmidrule(lr){3-5} \cmidrule(lr){6-7}
         & & \textbf{Random} & \textbf{Popular} & \textbf{Adver.} & $C_I$ & $C_S$ \\
        \midrule
        LLaVA-1.5-7B          & 1.0$\times$ & 83.29 & 81.88 & 78.96 & 15.9 & 52.8 \\
        \midrule
        + \textit{precomputed} & 1.0$\times$ & \textbf{89.87} & \textbf{88.32} & \textbf{84.27} & \textbf{11.5} & \textbf{34.6} \\
        + \textit{on the fly}  & 1.8$\times$ & 88.19 & 87.40 & 83.56 & 12.6 & 36.6 \\
        \bottomrule
    \end{tabular}
    
    \label{tab:latency_comparison}
\end{table}

\subsection{Distribution of Caption-Guided Attention Heads}
As illustrated in Figure \ref{fig:double}, we visualize the classification accuracies across 32 × 32 attention heads during the probing stage of LLaVA-1.5-7B (left) and Qwen-VL-Chat (right).
We observe that caption-guided attention heads are concentrated primarily between the 7th and 20th layers, which is well aligned with the layers with higher $\mathbf{Change\ Rate}$s presented in Figure \ref{fig:quan}. These attention heads play a critical role to fine-grained visual perception. By refining the output of these heads, CAST effectively enhances LVLM's visual perception capability and mitigates object hallucination.

\subsection{Impact of Hyperparameters and Inference Latency}

CAST method primarily relies on two key hyperparameters: the intensity of steering $\alpha$ and the number of refined attention heads $K$. We use grid search to find the optimal value for both hyperparameters on the POPE dataset.
See Appendix \ref{detailed_ablation} for detailed results. As shown in Figure \ref{fig:double}, we provide the ablation study results for each hyperparameter when the other is fixed to its optimal value. The key implications can be summarized as follows:

(1) Impact of $\alpha$: When $\alpha$ is small, the attention steering is insufficient, leading to marginal improvements. While a large $\alpha$ leads to insufficient attention to textual information, leading to a performance drop.

(2) Impact of $K$: Applying steering to few attention heads fails to influence the full activation pathways of visual information. While intervening in excess heads disrupts attention activation paths that are irrelevant to visual perception and play essential roles in other foundational capabilities, leading to performance drop.


Moreover, as shown in Table \ref{tab:chair}, we employ UniEval \citep{zhong2022towards} and perplexity (PPL) computation to evaluate the coherence and fluency of generated responses. We find even when doubling the optimal steering parameter, CAST does not compromise the coherence and fluency of outputs. Furthermore, as shown in Table \ref{latency}, CAST achieves better hallucination mitigating performance with less additional inference latency, which benefits from the inference-time steering paradigm.

\subsection{Discussion of CAST on The Fly}

Our default CAST implementation utilizes precomputed shift vectors to ensure both robustness and efficiency. By averaging attention differences over 1,000 samples, we extract a generalized direction for perceptual enhancement that filters out sample-specific semantic noise. Precomputation allows CAST to function as a plug-and-play module without introducing little additional inference-time cost.

To further explore the relationship between the extra computation required at inference time and the improvement achieved, we propose additional \textit{on-the-fly} approach. Concretely, we dynamically compute each inference sample's attention difference between the “caption query” and the “non-caption query” and employ this sample-specific vector for steering.
As shown in Table~\ref{tab:latency_comparison}, the \textit{on-the-fly} approach underperforms the precomputed version and incurs an 80\% increase in latency due to the requirement of two forward passes. These results suggest that the precomputed vector offers a more stable and efficient optimization direction, making it the superior choice for practical applications.

\subsection{Case Study}

CAST remains effective in caption task, which is attributed to the enhancement in visual attention. As shown in Figure \ref{fig:case_study}, CAST effectively mitigates object hallucination not only during the regeneration of new responses, but also when extending hallucinated contexts, highlighting its fine-grained, token-level object hallucination mitigation capability.



\section{Conclusion}
In this paper, we are the first to explicitly reveal the impact of caption queries versus non-caption queries on the attention activation patterns of LVLMs, providing novel insights for the optimization of visual attention. Furthermore, we propose CAST, a training-free method that probes and refines caption-guided attention heads during inference, thereby enhancing LVLM's fine-grained perception capabilities and mitigating object hallucination. 
Comprehensive experimental results across five widely used benchmarks demonstrate that CAST not only effectively mitigates hallucination with little inference latency, but also shows strong generalization.

\section*{Impact Statement}
This paper presents work whose goal is to advance the field of machine learning. There are many potential societal consequences of our work, none of which we feel must be specifically highlighted here.

\bibliography{example_paper}
\bibliographystyle{icml2026}

\newpage
\appendix
\onecolumn

\section*{Appendix Contents}
\vspace{0.2em}

{\small
\setlength{\parindent}{0pt}
\setlength{\parskip}{0.5em}      
\renewcommand{\arraystretch}{1.15}

\newcommand{\tocI}[2]{\hyperref[#1]{\textbf{\ref*{#1}.} #2}\par}
\newcommand{\tocII}[2]{\hspace*{1.9em}\hyperref[#1]{\ref*{#1}. #2}\par}

\tocI{sec:B}{Experimental Setup of Quantitative Analysis}
\tocI{sec:C}{Additional Experimental Details}
\tocII{benchmarks}{Benchmarks}
\tocII{sec:data_source}{Data Source}
\tocII{sec:detail_setup}{Detailed Experimental Setup}
\tocI{sec:more_methods}{Comparison with More Advanced Methods}
\tocI{ad_models}{Results on More Advanced Models}
\tocI{more_bench}{Results on More Advanced Benchmarks}
\tocI{sec:F}{Results on Domain-Specific Benchmarks}
\tocI{layers}{Discussion on the CAST Steering Layers}
\tocI{sec:impacts}{Impacts of the Classifier Types and Training Data}
\tocII{sec:impacts_cls}{Impacts of the Classifier Types}
\tocII{sec:impacts_data}{Impacts of the Classifier Training Data}
\tocI{detailed_ablation}{Fine-grained Analysis of Hyperparameters}
\tocII{ab_pope}{Grid-search Results of LLaVA-1.5-7B on POPE}
\tocII{ab_chair}{Grid-search Results of LLaVA-1.5-7B on CHAIR}
\tocI{sec:E}{Detailed Experimental Results of MME}
\tocI{evidence}{Deeper Evidence of CAST’s Effectiveness in Discriminative Settings}
\tocI{sec:case_cap}{Case Study for Caption Queries}
\tocI{sec:H}{Case Study for Non-caption Queries}
\tocI{sec:llm_usage}{Usage of Large Language Models}
}

\vspace{0.4em}
\hrule
\vspace{0.8em}

\section{Experimental Setup of Quantitative Analysis}
\label{sec:B}
We sample 1,000 images from the MS-COCO dataset \citep{lin2014microsoft}. For each image, we propose one caption query and two different non-caption queries (non-caption-1 \& non-caption-2) to analyze differences attributable to query types. 

We consider a LVLM parametrized by \( \theta \). The model receives as input a textual query $ \boldsymbol{T} = \{t_1, t_2, \dots, t_n\} $ and a visual input $ \boldsymbol{V} = \{v_1, v_2, \dots, v_m\} $, where $ n $ and $ m $ denote the sequence lengths of the text and visual inputs. The text and vision inputs are concatenated together to form the first layer input $ \boldsymbol{H}^{1} = \mathrm{concat}(\boldsymbol{V},\boldsymbol{T}) \in \mathbb{R}^{(m+n)\times d}$ for the $L$ layers $\times$ $H$ heads decoder. For an image, the last input token's visual attention weight of $H$-th head in $L$-th layer $\boldsymbol{Sum}_{(l,h)}$ can be computed as:  

\begin{equation}
\boldsymbol{A}_{(l,h)} = \textrm{softmax}( \frac{\boldsymbol{Q}_{(l,h)} \boldsymbol{K}_{(l,h)}^T}{\sqrt{d}}),
\end{equation}

\begin{equation}
\boldsymbol{Sum}_{(l,h)}=\sum_{i=1}^{m}\boldsymbol{A}^{-1}_{(l,h)}[i],
\end{equation}

where the $\boldsymbol{Q}_{(l,h)}$ and $\boldsymbol{K}_{(l,h)}$ are the Query and Key matrixs of the $k$-th head in $l$-th layer, $\boldsymbol{A}^{-1}_{(l,h)}[i]$ is the last input token's attention weight of the $i$-th input token. For a dataset of $B$ samples, the sum of visual attention weight can be computed as:

\begin{equation}
{S}_{(l,h)}=\sum_{b=1}^{B}\boldsymbol{Sum}_{(l,h)}.
\end{equation}

Then we record the sum of visual attention weights from the last input token for three types of queries: $S^{cap}_{(l,h)}$ for caption query, $S^{non-1}_{(l,h)}$ for non-caption query 1 and $S^{non-2}_{(l,h)}$ for non-caption query 2. The head-wise Change Rate $Rate_{(l,h)}$ and layer-wise Change Rate $Rate_{(l)}$ can be computed as:

\begin{equation}
Rate_{(l,h)}^{cap}=\frac{S^{cap}_{(l,h)} - S^{non-1}_{(l,h)}}{S^{non-1}_{(l,h)}}, Rate_{(l,h)}^{non-cap}=\frac{S^{non-2}_{(l,h)} - S^{non-1}_{(l,h)}}{S^{non-1}_{(l,h)}},
\end{equation}

\begin{equation}
    Rate_{(l)}^{cap}=\frac{\sum_{h=1}^{H}{(S^{cap}_{(l,h)} - S^{non-1}_{(l,h)})}}{\sum_{h=1}^{H}S^{non-1}_{(l,h)}},
    Rate_{(l)}^{non-cap}=\frac{\sum_{h=1}^{H}{(S^{non-2}_{(l,h)} - S^{non-1}_{(l,h)})}}{\sum_{h=1}^{H}S^{non-1}_{(l,h)}}.
\end{equation}

By comparison, we find that visual attention across particular attention heads was significantly enhanced when fed caption compared to non-caption queries. These results provide strong support for our proposed motivation.

\section{Additional Experimental Details}
\label{sec:C}
\subsection{Benchmarks}
\label{benchmarks}
We evaluate our proposed CAST method across five benchmarks, including both discriminative and generative tasks to measure its effectiveness and robustness:

\textbf{(1) POPE} \citep{li2023evaluating} employs a binary question-answering format, inquiring LVLMs to answer if a special object exists in the given image. We adopt Accuracy and F1 score as the evaluation metrics.

\textbf{(2) MME} \citep{Fu2023MMEAC} serves as a comprehensive tool for assessing the capabilities of LVLMs across 10 perception tasks and 4 cognition tasks. 
Consequently, task scores are reported as the evaluation metrics.

\textbf{(3) CHAIR} \citep{rohrbach2018object} is a widely used metric to assess object hallucination of LVLMs. 
The CHAIR metric comprises two indicators, denoted as $C_S$ and $C_I$, with the following calculation formulas:
    \begin{equation}
\small
   C_S = \frac{|\{\textrm{Hallucinated}\ \textrm{objects}\}|}{|\{\textrm{All}\ \textrm{mentioned}\ \textrm{objects}\}|}  \notag
\end{equation}
\begin{equation}
    \small
 C_I = \frac{|\{\textrm{Sentences w/ hallucinated objects}\}|}{|\{\textrm{All sentences}\}|} \notag
\end{equation}

\textbf{(4) MMHal-Bench} \citep{sun2023aligning} comprises 96 meticulously designed questions, which evaluates response-level hallucination rate (VH.\%) and informativeness (Score). It asks \textbf{GPT-4} \citep{achiam2023gpt} to compare model outputs with human responses and object labels for evaluation.

\textbf{(5) MHumanEval} \citep{yu2024rlaif} is designed to evaluate hallucination performance by \textbf{human annotators}. The benchmark contains 146 samples collected from Object HalBench and MMHal-Bench. Given model responses, we ask three human annotators to label the hallucinated segments and compute the mean response-level hallucination rate (Hu.\%) as the evaluation metric.
\subsection{Data Source}
\label{sec:data_source}
Although our method does not rely on specific data, we separately specify the sources of the data used in the experiments for the sake of reproducibility.
\subsubsection{Data of Best Query Search}
In the best caption search algorithm, we use the top 100 VQA samples from the complex reasoning data in the LLaVA-1.5-7b pre-training dataset. From this, we obtain non-caption queries and their corresponding images. Additionally, we maintain a list of 16 candidate caption queries, some of which are manually generated and others are derived from the pre-trained instructions of LLaVA-1.5-7b. The caption query candidates are listed as follows:

\textit{"What do you see happening in this image?"}, 
\textit{"What do you think is going on in this snapshot?"},
\textit{"Can you elaborate on the elements of the picture provided?"},
\textit{"Describe the following image."},
\textit{"What's happening in the scene?"},
\textit{"Analyze the image in a comprehensive and detailed manner."},
\textit{"Write a detailed description of the given image."},
\textit{"What is this photo about?"},
\textit{"Explain the visual content of the image in great detail."},
\textit{"What are the key elements in this picture?"},
\textit{"Can you describe the main features of this image for me?"},
\textit{"Please describe this image in detail."},
\textit{"Generate the caption in English:"}
\textit{"Provide a thorough narrative of what the image depicts."}
\textit{"Offer a detailed explanation of the scene captured in the picture."}
\textit{"Summarize the visual information conveyed by this image."}

In the experiments, the best caption query for LLaVA-1.5-7b and LLaVA-NeXT is \textit{"Analyze the image in a comprehensive and detailed manner."} and the best caption query for Qwen-VL-Chat, InternVL2-8B, Qwen2-VL-7B and Qwen2.5-VL-7B is \textit{"Please describe this image in detail."}.

\subsubsection{Data of Probe and Shift Computation }
We extracted the first 1,000 samples from the complex reasoning data in the LLaVA-1.5-7b pre-training dataset. The questions from these samples were treated as non-caption queries.
\subsection{Detailed Experimental Setup}
\label{sec:detail_setup}
In the experiment of POPE, 'regular' refers to the direct sampling setting. We used direct sampling decoding and set $\alpha=1.5$ and $K = 100$ in the main experiments.

\section{Comparison with More Advanced Methods}
\label{sec:more_methods}
We selected LLaVA-1.5-7b as the baseline model and compared CAST with more advanced models including VCD \citep{leng2024mitigating}, ICD \citep{wang2024mitigating}, OPERA \citep{huang2024opera}, Woodpecker \citep{yin2024woodpecker}, M3ID \citep{favero2024multi}, DAMRO \citep{gong2024damro}, IMCCD \citep{li2025mitigating}, CATCH \citep{kan2024catch}, IBD \citep{zhu2024ibd}, CAUSALMM \citep{zhou2024mitigating} and ICT \citep{chen2025ict}.
The results of CAST compared with more SOTA methods on MS-COCO POPE are shown in Table \ref{add_coco_result}.

\begin{table*}[!ht]
\small
\centering
\renewcommand{\arraystretch}{1.0} 
\setlength{\tabcolsep}{1.7pt} 
\caption{Result compared with more advanced methods on MS-COCO POPE. 
}
\begin{tabular}{lcccccccc}
\toprule
\multirow{2}{*}{ \textbf{Method}} & \multicolumn{2}{c}{\textbf{Random}} & \multicolumn{2}{c}{\textbf{Popular}} & \multicolumn{2}{c}{\textbf{Adversarial}} & \multicolumn{2}{c}{\textbf{Average}}  \\
\cmidrule(lr){2-3}
\cmidrule(lr){4-5}
\cmidrule(lr){6-7}
\cmidrule(lr){8-9}
& Accuracy & F1-Score & Accuracy & F1-Score & Accuracy & F1-Score & Accuracy & F1-Score \\
\midrule
Regular &83.29  &81.33  &81.88  &80.06 &78.96 &77.57&81.38&79.65 \\
ICD \textcolor{gray}{\textit{(EMNLP'24 findings)}}&89.56 &89.68 &86.16&86.76 &79.71 &81.70 &85.14 &86.05\\
OPERA \textcolor{gray}{\textit{(CVPR'24)}} &89.20  &88.81  &86.64  &86.62 &81.24 &81.38& 85.70& 85.60\\
Woodpecker \textcolor{gray}{\textit{(SCIS'24)}}&87.67 &86.45 &80.67 &79.72 &80.67 &80.00&83.00&82.05 \\
M3ID \textcolor{gray}{\textit{(CVPR'24)}} &86.20 &84.51 & 84.77& 83.17& 82.53 &81.14 &84.50&82.94 \\
DAMRO \textcolor{gray}{\textit{(EMNLP'24)}}&88.20 &87.29 & 85.67& 84.98 &82.07&81.90&85.31&84.72\\
IMCCD \textcolor{gray}{\textit{(arXiv'25)}}& 89.23 &88.68 &86.73 &86.13 &82.87 & 82.77&86.27&85.86\\
CATCH \textcolor{gray}{\textit{(ECCV'24)}} & \textbf{90.43} & \textbf{90.13} &87.07 &86.56 & 83.17& 83.18& 86.89 &86.62\\
VDD \textcolor{gray}{\textit{(arXiv'24)}}& 90.00 & 88.79 & 85.91& 84.40 & 83.52 &82.20&86.48&85.13\\
CAUSALMM \textcolor{gray}{\textit{(ICLR'25)}} & 88.93 &88.10 & 87.13 & 87.26 &83.70 &82.78&86.59&86.05 \\ 
ICT \textcolor{gray}{\textit{(CVPR'25)}} &90.11&90.03&87.50&87.60&\textbf{84.43}&83.74&87.35&87.12\\
\midrule
CAST(ours) & 89.87 & 89.43 &  \textbf{88.32} & \textbf{87.95} &84.27 & \textbf{84.41} &\textbf{87.49} &\textbf{87.22}\\
\bottomrule
\end{tabular}
\label{add_coco_result}
\end{table*}

To further demonstrate the superiority of CAST's performance, we additionally compare CAST with two advanced RL methods, including HADPO \citep{zhao2023hallucinations} and HALVA \citep{sarkar2024mitigating}. As shown in the Table \ref{tab:pope_chair_mme} and Table \ref{tab:hallusion_gavie}, CAST achieves performance comparable to these RL methods and even surpasses them on discriminative tasks.

\begin{table*}[!ht]
\caption{Comparisons between CAST and RL works on POPE, CHAIR, and MME benchmarks.}
\centering
\small
\resizebox{\textwidth}{!}{%
\begin{tabular}{lccccccccc}
\toprule
\multirow{2}{*}{\textbf{Method}} & \multicolumn{3}{c}{\textbf{POPE}} & \multicolumn{2}{c}{\textbf{CHAIR} ($\downarrow$)} & \multicolumn{4}{c}{\textbf{MME}} \\
\cmidrule(lr){2-4} \cmidrule(lr){5-6} \cmidrule(lr){7-10}
 & \textbf{Random} & \textbf{Popular} & \textbf{Adver.} & \textbf{CHAIR}$_I$ & \textbf{CHAIR}$_S$ & \textbf{Count} & \textbf{Exist.} & \textbf{Color} & \textbf{Posi.} \\
\midrule
LLaVA-1.5-7B & 83.29 & 81.88 & 78.96 & 15.9 & 52.8 & 124.67 & 175.67 & 151.00 & 114.00 \\
+ HADPO      & 86.00 & 85.10 & 82.90 & \textbf{11.0} & 38.2 & 133.30 & \textbf{190.00} & 158.30 & 136.70 \\
+ HALVA      & 86.40 & 85.50 & 83.20 & 11.7 & 41.4 & \textbf{165.00} & \textbf{190.00} & \textbf{175.00} & 135.00 \\
+ CAST        & \textbf{89.87} & \textbf{88.32} & \textbf{84.27} & 11.5 & \textbf{34.6} & 141.67 & \textbf{190.00} & 170.00 & \textbf{140.00} \\
\bottomrule
\end{tabular}%
}
\label{tab:pope_chair_mme}
\end{table*}

\begin{table*}[!ht]
\centering
\small 
\setlength{\tabcolsep}{4pt} 
\caption{Comparisons between CAST and RL works on HallusionBench and GAVIE benchmarks.}
\begin{tabular}{lccccccc}
\toprule
\multirow{2}{*}{\textbf{Method}} & \multicolumn{5}{c}{\textbf{HallusionBench}} & \multicolumn{2}{c}{\textbf{GAVIE}} \\
\cmidrule(lr){2-6} \cmidrule(lr){7-8}
 & \textbf{qAcc} & \textbf{fAcc} & \textbf{Easy aAcc} & \textbf{Hard aAcc} & \textbf{aAcc} & \textbf{Relevancy} & \textbf{Accuracy} \\
\midrule
LLaVA-1.5-7B & 10.55 & 20.86 & 41.67 & 29.77 & 46.04 & 8.20 & 6.42 \\
+ HADPO      & 11.21 & 19.08 & 42.86 & 39.19 & 47.46 & \textbf{8.84} & 6.30 \\
+ HALVA      & \textbf{13.85} & \textbf{21.48} & 42.71 & \textbf{40.81} & \textbf{47.95} & 8.72 & 6.46 \\
+ CAST        & 12.90 & 20.96 & \textbf{43.34} & 37.69 & 46.75 & 8.76 & \textbf{6.68} \\
\bottomrule
\end{tabular}
\label{tab:hallusion_gavie}
\end{table*}

\section{Results on More Advanced Models}

\label{ad_models}

As shown in Table \ref{advancemodel}, CAST further exhibits effective hallucination mitigation when applied to more advanced models, providing additional evidence for the generalizability of CAST.

\begin{table}[!ht] 
\small
\centering
\renewcommand{\arraystretch}{1} 
\setlength{\tabcolsep}{3pt} 
\caption{Results on more advanced LVLMs, including Qwen2-VL-7B \citep{Qwen2-VL}, InternVL2-8B \citep{chen2024far}, LLaVA-NeXT and Qwen2.5-VL-7B \citep{bai2025qwen2}. Cog. and Hall. denote the cognitive and hallucination subset of MME benchmark.
}
\begin{tabular}{rcccccccccc}
\toprule
\multirow{2}{*}{ \textbf{Model\ \ \ \ \ \ }} & \multicolumn{2}{c}{\textbf{POPE}} & \multicolumn{2}{c}{\textbf{MME}} & \multicolumn{2}{c}{\textbf{CHAIR}} & \multicolumn{3}{c}{\textbf{MMHal-Bench}} \\
\cmidrule(lr){2-3}
\cmidrule(lr){4-5}
\cmidrule(lr){6-7}
\cmidrule(lr){8-10}
&Acc(\%) \textbf{$\uparrow$} & F1-Score(\%) \textbf{$\uparrow$}
& Cog.\textbf{$\uparrow$} & Hall.\%\textbf{$\uparrow$} & $C_S$ \textbf{$\downarrow$} &  $C_I$ \textbf{$\downarrow$} &Score\textbf{$\uparrow$} & VH.\%\textbf{$\downarrow$} & Hu.\%\textbf{$\downarrow$} \\
\midrule
Qwen2-VL-7B & 88.49 & 87.85 & 556.4 & 630.0& 24.8&7.2 &2.87 &49.8 &55.4 \\
\textbf{+ CAST} & \textbf{89.85} & \textbf{89.87} &  \textbf{570.4} & \textbf{668.3}& \textbf{15.6}&\textbf{6.5} & \textbf{3.09} &\textbf{40.8} &\textbf{48.4}\\
\midrule
InternVL2-8B & 86.67 & 85.72 & 566.4 & 663.0 & 37.2&9.4 &2.71 &52.3&56.7\\
\textbf{+ CAST} & \textbf{87.98} & \textbf{87.42} &  \textbf{573.3} & \textbf{693.7} & \textbf{31.3} &\textbf{8.4} & \textbf{2.91} &\textbf{44.4}&\textbf{49.7}\\
\midrule
LLaVA-NeXT & 83.06  &84.57&533.7&586.7&40.0&10.5&2.57&55.8&65.4 \\
\textbf{+ CAST} & \textbf{88.73} &\textbf{88.58}&\textbf{566.7}&\textbf{657.5}&\textbf{33.3}&\textbf{8.9}&\textbf{3.12}&\textbf{48.9}&\textbf{61.0} \\
\midrule
Qwen2.5-VL-7B &87.35 &87.09 &630.0 &683.3 &37.2 &8.7 &3.05 &34.7 &43.6 \\
\textbf{+ CAST} &\textbf{88.96} &\textbf{88.70} &\textbf{655.7} &\textbf{695.0} &\textbf{32.6} &\textbf{8.0} &\textbf{3.24} &\textbf{29.9} &\textbf{40.2}\\

\bottomrule
\end{tabular}
\label{advancemodel}
\end{table}









\section{Results on More Advanced Benchmarks}
\label{more_bench}
The five commonly used hallucination evaluation benchmarks included in our paper follow the setups adopted in recent works. Using these benchmarks allows us to make fair and comprehensive comparisons with prior training-free methods.
Furthermore, we additionally conduct experiments on more advanced hallucination evaluation benchmarks, including HallusionBench \citep{wu2024autohallusion} and GAVIE \citep{liu2023aligning}. As shown in Table \ref{tab:hallusion_gavie_all}, CAST also achieves improvements on these more critical evaluation.

\begin{table*}[!ht]
\centering

\small
\setlength{\tabcolsep}{9pt}
\caption{Comparisons on HallusionBench and GAVIE benchmarks across different MLLMs.}
\begin{tabular}{lccccccc}
\toprule
\multirow{2}{*}{\textbf{Method}} & \multicolumn{5}{c}{\textbf{HallusionBench}} & \multicolumn{2}{c}{\textbf{GAVIE}} \\
\cmidrule(lr){2-6} \cmidrule(lr){7-8}
 & \textbf{qAcc} & \textbf{fAcc} & \textbf{Easy aAcc} & \textbf{Hard aAcc} & \textbf{aAcc} & \textbf{Relevancy} & \textbf{Accuracy} \\
\midrule
LLaVA-1.5-7B  & 10.55 & 20.86 & 41.67 & 29.77 & 46.04 & 8.20 & 6.42 \\
+ CAST         & \textbf{12.90} & \textbf{20.96} & \textbf{43.34} & \textbf{37.69} & \textbf{46.75} & \textbf{8.76} & \textbf{6.68} \\
\midrule
Qwen-VL-Chat  & 8.93 & 11.56 & 34.43 & 28.87 & 41.12 & 8.26 & 6.39 \\
+ CAST         & \textbf{11.47} & \textbf{13.57} & \textbf{35.60} & \textbf{31.87} & \textbf{43.93} & \textbf{8.63} & \textbf{6.60} \\
\midrule
Qwen2.5-VL-7B & 16.43 & 31.01 & 59.73 & 34.93 & 50.79 & 9.20 & 8.09 \\
+ CAST         & \textbf{19.73} & \textbf{32.31} & \textbf{64.56} & \textbf{45.40} & \textbf{53.80} & \textbf{9.33} & \textbf{8.42} \\
\bottomrule
\end{tabular}
\label{tab:hallusion_gavie_all}
\end{table*}

\section{Results on Domain-Specific Benchmarks}
\label{sec:F}

In domain-specific tasks, the CAST method demonstrates certain generalization ability to some extent. Although caption queries are general instructions, they are extensively used during model pretraining. Activating the relevant attention patterns facilitates fine-grained visual information capture, thereby enhancing downstream task performance. To evaluate CAST's effectiveness in specific domains, we selected VQA-RAD \citep{lau2018dataset} from the medical domain and the MMBench \citep{liu2024mmbench} OCR subset. The experimental results of LLaVA-1.5-7b, as presented in the table \ref{domain}, show consistent improvements over the baseline, indicating the CAST method's generalization ability.

\begin{table}[!ht]
\small
\centering
\renewcommand{\arraystretch}{1} 
\setlength{\tabcolsep}{6pt} 
\caption{Results on VQA-RAD and MMbench OCR subset.}.
\begin{tabular}{lccc}
\toprule
Domain & Dataset & Method & Accuracy  \\
\midrule
Medical & VQA-RAD & Greedy & 54.18\% \\
&&CAST & 58.17\%\\
\midrule
OCR & MMBench& Greedy & 74.31\%\\
&& CAST & 77.54\% \\
\bottomrule
\end{tabular} 
\label{domain}
\end{table}

\section{Discussion on the CAST Steering Layers}
\label{layers}
CAST method adds steering across all model layers rather than targeting in a certain layer, based on prior studies on information flow \citep{li2025causal,golovanevsky2024vlms,neo2024towards,meng2022locating}, we argue that steering on attention heads in a single layer alone cannot effectively enhance visual perception; these important attention heads must be activated or perturbed across layers to fully reinforce the visual information flow \citep{neo2024towards,meng2022locating}. Steering only in shallow layers without affecting higher layers may impair perception, while steering only in higher layers cannot fully strengthen the visual processing information flow \citep{li2025causal}, limiting CAST’s ability to achieve optimal hallucination mitigation. As shown in Table ~\ref{tab:layer_ablation}, our experiments further confirm this: steering on top-100 caption-guided heads in layers 0–10, 11–20 and 21-31 alone does not achieve optimal CAST performance and may even degrade model capability.

\begin{table}[!ht]
    \centering
    \small 
    \setlength{\tabcolsep}{4pt} 
    \caption{\textbf{Ablation study on steering layers.} We apply CAST to different blocks of layers to identify the most critical stages. The results show that steering in the middle layers (11-20) yields more significant improvements than early or late layers, while the full CAST method achieves the best performance by coordinating across all identified heads.}    
    \begin{tabular}{lccccc}
        \toprule
        \multirow{2}{*}{\textbf{Method}} & \multicolumn{3}{c}{\textbf{POPE} ($\uparrow$)} & \multicolumn{2}{c}{\textbf{CHAIR} ($\downarrow$)} \\
        \cmidrule(lr){2-4} \cmidrule(lr){5-6}
         & \textbf{Random} & \textbf{Popular} & \textbf{Adver.} & $C_I$ & $C_S$ \\
        \midrule
        LLaVA-1.5-7B   & 83.29 & 81.88 & 78.96 & 15.9 & 52.8 \\
        + CAST w/ 0-10  & 82.07 & 80.65 & 77.41 & 16.4 & 54.0 \\
        + CAST w/ 11-20 & 87.16 & 85.83 & 82.52 & 13.0 & 38.2 \\
        + CAST w/ 21-31 & 86.78 & 84.22 & 80.87 & 15.4 & 44.3 \\
        \midrule
        + CAST (\textit{Ours}) & \textbf{89.87} & \textbf{88.32} & \textbf{84.27} & \textbf{11.5} & \textbf{34.6} \\
        \bottomrule
    \end{tabular}
    
    \label{tab:layer_ablation}
\end{table}

\section{Impacts of the Classifier Types and Training Data}
\label{sec:impacts}
\subsection{Impacts of the Classifier Types}
\label{sec:impacts_cls}
Inspired by prior works \citep{li2023inference,bao2025probing,zhang2024pip}, which show that SVM effectively performs binary classification on high-dimensional internal model vectors, we adopt SVM as the classifier in our CAST framework.
To further analyze the impacts of the classifier types, we implement \textbf{Logistic Regression (LR)} as an alternative classifier. The experimental results are shown in the table below. CAST with LR achieves performance nearly identical to CAST with SVM, as 95\% of the Top-100 attention heads selected by both classifiers are the same. CAST with SVM exhibits a slight performance advantage, which aligns with findings in related work \citep{wang2025thoughtprobe} and further confirms SVM’s superior capability in classifying high-dimensional vectors.

\begin{table*}[!ht]
\centering
\caption{Performance comparison between SVM and LR classifiers on POPE, CHAIR, and MME benchmarks. The best results are highlighted in \textbf{bold}.}
\label{tab:main_results}
\resizebox{\textwidth}{!}{%
\begin{tabular}{lccccccccc}
\toprule
\multirow{2}{*}{\textbf{Model}} & \multicolumn{3}{c}{\textbf{POPE} ($\uparrow$)} & \multicolumn{2}{c}{\textbf{CHAIR} ($\downarrow$)} & \multicolumn{4}{c}{\textbf{MME} ($\uparrow$)} \\
\cmidrule(lr){2-4} \cmidrule(lr){5-6} \cmidrule(lr){7-10}
 & \textbf{Random} & \textbf{Popular} & \textbf{Adversarial} & $C_I$ & $C_S$& \textbf{Count} & \textbf{Exist.} & \textbf{Color} & \textbf{Posi.} \\
\midrule
LLaVA-1.5-7b & 83.29 & 81.88 & 78.96 & 15.9 & 52.8 & 124.67 & 175.67 & 151.00 & 114.00 \\
\midrule
+ CAST w/ SVM & \textbf{89.87} & \textbf{88.32} & \textbf{84.27} & \textbf{11.5} & \textbf{34.6} & \textbf{141.67} & \textbf{190.00} & \textbf{170.00} & \textbf{140.00} \\
+ CAST w/ LR  & 89.40 & 88.13 & 83.87 & 11.7 & 34.9 & 138.33 & \textbf{190.00} & \textbf{170.00} & 135.00 \\
\bottomrule
\end{tabular}%
}
\end{table*}

\subsection{Impacts of the Classifier Training Data}
\label{sec:impacts_data}
To further investigate the amount of classifier training data, we randomly select distinct samples from the whole LLaVA-1.5-7B pre-training dataset (77K) and retrain the classifiers. We evaluate the classifying consistency of Top-$k$ heads using the \textbf{Overlap Ratio}, defined as $|H_{n} \cap H_{CAST}|/k$, where $n$ is the number of samples, $n \in \{100,250,500,1500,2000,5000\}$; $k\in\{50,100\}$; $H_{n}$ denotes heads identified by new classifiers and $H_{CAST}$ denotes heads identified in our primary results. The following table shows that the classifier's training is robust to data variations and amount, as the top-100 caption-guided attention heads which play a critical role in visual perception \textbf{predominantly coincide with} the CAST identified in the paper.

\begin{table}[htbp]
    \centering
    \caption{Robustness analysis of classifier training. The high overlap ratios across varying sample sizes ($n$) and Top-$k$ attention heads demonstrate that the identified attention heads are consistent and robust to data amount variations compared to the primary setting ($n=1000$).}
    \resizebox{\textwidth}{!}{%
        \begin{tabular}{lccccccc}
            \toprule
            \textbf{Overlap Ratio} & $n=100$ & $n=250$ & $n=500$ & \textbf{$n=1000$ (CAST)} & $n=1500$ & $n=2000$ & $n=5000$ \\
            \midrule
            $k=10$  & 0.90 & 1.00 & 1.00 & 1.00 & 1.00 & 1.00 & 1.00 \\
            $k=50$  & 0.94 & 0.96 & 0.96 & 1.00 & 0.98 & 1.00 & 0.98 \\
            $k=100$ & 0.88 & 0.90 & 0.93 & 1.00 & 0.95 & 0.94 & 0.94 \\
            \bottomrule
        \end{tabular}%
    }
    
    \label{tab:data_robustness}
\end{table}

\section{Fine-grained Analysis of Hyperparameters}
\label{detailed_ablation}
\subsection{Grid-search Results of LLaVA-1.5-7B on POPE}
\label{ab_pope}
In our experiments, we conducted a grid search to identify the optimal values of the hyperparameters $\alpha$ and $K$. We now provide the full grid-search results of LLaVA-1.5-7B on the POPE Adversarial benchmark, which gives a more clear and continuous view of how $\alpha$ and $K$ jointly affect CAST’s performance. The best performance is achieved at $\alpha=1.5$ and $K=100$.
In our main experiments, we applied this parameter to the evaluation of all test sets and consistently achieved significant performance improvements.

\begin{table}[!ht]
    \centering
    \small  
    \setlength{\tabcolsep}{5pt} 
    \renewcommand{\arraystretch}{1.0} 
    \caption{Grid-search results on POPE-Adversarial.
    }
    \begin{tabular}{lccccccc}
        \toprule
        \textbf{Accuracy} 
         & $\alpha=0$ & $\alpha=0.5$ & $\alpha=1.0$ & $\alpha=1.25$ & $\alpha=1.5$ & $\alpha=1.75$ & $\alpha=2.0$ \\ 
        \midrule
        $K=0$   & 78.96 & 78.96 & 78.96 & 78.96 & 78.96 & 78.96 & 78.96 \\
        $K=50$  & 78.96 & 79.31 & 79.86 & 80.18 & 80.50 & 80.32 & 80.40 \\
        $K=75$  & 78.96 & 79.82 & 80.13 & 80.44 & 80.77 & 80.59 & 80.31 \\
        $K=100$ & 78.96 & 81.07 & 82.50 & 83.47 & \textbf{84.27} & 84.14 & 84.00 \\
        $K=125$ & 78.96 & 80.79 & 82.16 & 83.28 & 84.10 & 83.82 & 83.51 \\
        $K=150$ & 78.96 & 80.24 & 81.47 & 82.53 & 83.20 & 82.97 & 82.68 \\
        $K=200$ & 78.96 & 79.91 & 81.18 & 82.12 & 83.00 & 82.76 & 82.43 \\
        \bottomrule
    \end{tabular}

    \label{tab:grid_search}
\end{table}

\subsection{Grid-search Results of LLaVA-1.5-7B on CHAIR}
\label{ab_chair}
CAST can achieve slightly better performance with task-specific hyperparameters in some generative tasks. As shown in the table, we conducted hyperparameter analysis on the CHAIR benchmark. The optimal parameters are found to be ($\alpha = 1.25$, $K = 125$ and performance = 34.3); nevertheless, the performance difference compared to the POPE-optimal parameters ($\alpha = 1.5$, $K = 100$ and performance = \textit{34.6}) is minimal.

\begin{table}[!ht]
    \centering
    \small 
    \setlength{\tabcolsep}{5pt} 
    \renewcommand{\arraystretch}{1.0} 
    \caption{Grid-search results on CHAIR.}
    \begin{tabular}{lcccccc}
        \toprule
         $C_S\downarrow$
         & $\alpha=0$ & $\alpha=1.0$ & $\alpha=1.25$ & $\alpha=1.5$ & $\alpha=1.75$ & $\alpha=2.0$ \\
        \midrule
        $K=0$   & 52.8 & 52.8 & 52.8 & 52.8 & 52.8 & 52.8 \\
        $K=50$  & 52.8 & 44.3 & 43.1 & 43.5 & 44.0 & 44.8 \\
        $K=75$  & 52.8 & 39.6 & 37.5 & 37.6 & 38.6 & 39.4 \\
        $K=100$ & 52.8 & 35.1 & 34.4 & \textit{34.6} & 35.2 & 35.9 \\
        $K=125$ & 52.8 & 34.9 & \textbf{34.3} & 34.5 & 35.0 & 35.7 \\
        $K=150$ & 52.8 & 35.3 & 34.7 & 35.1 & 35.8 & 36.5 \\
        $K=200$ & 52.8 & 36.0 & 34.4 & 36.1 & 36.7 & 37.3 \\
        \bottomrule
    \end{tabular}
    
    \label{tab:chair_grid_search}
\end{table}

Nevertheless, we observe that the optimal parameters identified on POPE Adversarial dataset \textbf{can generalize well to other discriminative and generative tasks} (e.g., MME, CHAIR, MMHal-Bench). This indicates that the fixed optimal hyperparameters can be effectively applied in real-world scenarios, demonstrating \textbf{CAST’s ease of deployment and strong generalization capability}.

\section{Detailed Experimental Results of MME}
\label{sec:E}
Detailed results on MME perception and cognition can be found in Table \ref{add_mme_p_result} and Table \ref{add_mme_r_result}.

\begin{table*}[!ht]
\small
\centering
\renewcommand{\arraystretch}{1.1} 
\setlength{\tabcolsep}{1.5pt} 
\caption{Results on all MME perception-related tasks. The best performance of each is \textbf{bolded}. 
}
\begin{tabular}{lccccccccccc}
\toprule
Method & Artwork &Celebrity &Color &Count &Existence & Landmark &OCR &Position &Posters &Scene & Total \\
\midrule
Regular &102.20 &113.59 &151.00 &124.67 &175.67&129.95 &92.00 &114.00 &127.82 &148.30 &1279.20\\
VCD & 109.60&120.94&153.00&138.33&184.66&140.45&104.00&128.67&132.11&152.20&1363.96\\
OPERA &\textbf{122.50}&126.76&149.00&116.00&165.00&152.75&\textbf{132.50}&133.33&136.05&154.00&1387.89 \\
CAST(ours)&120.25 &\textbf{135.88} &\textbf{170.00} &\textbf{141.67}& \textbf{190.00}&\textbf{158.50}&120.00&\textbf{140.00} &\textbf{140.48}&\textbf{157.00}&\textbf{1473.78} \\
\bottomrule
\end{tabular}
\label{add_mme_p_result}
\end{table*}

\begin{table*}[!ht]
\small
\centering
\renewcommand{\arraystretch}{1.1} 
\setlength{\tabcolsep}{2.5pt} 
\caption{Results on all MME recognition-related tasks. The best performance is \textbf{bolded}. 
}
\begin{tabular}{lccccc}
\toprule
Method & Coding Reasoning &Commonsense Reasoning & Numerical Calculation& Text Translation& Total \\
\midrule
Regular &66.38 &106.43&57.00&72.50&302.31 \\
VCD &68.50&111.29&42.64&68.50 &290.93 \\
OPERA &62.50 &119.29 &37.50 &\textbf{82.50}&301.79 \\
CAST(ours)& \textbf{75.00} &\textbf{122.86} &\textbf{57.50} &80.00&\textbf{335.36} \\
\bottomrule
\end{tabular}
\label{add_mme_r_result}
\end{table*}

\section{Deeper Evidence of CAST’s Effectiveness in Discriminative Settings}
\label{evidence}
Previous works \citep{sarkar2024mitigating,liu2023mitigating} observed that the “yes-bias” in discriminative tasks arises because "models are finetuned on unbalanced datasets containing predominantly positive instructions" \citep{liu2023mitigating}, and thus represents the main form of LVLM's object hallucination. Furthermore, we computed the confusion matrices of LLaVA-1.5-7B on the POPE popular and random subsets. As shown in the Table, CAST substantially mitigates the “yes-bias,” providing deeper evidence of CAST’s effectiveness in discriminative settings.

\begin{table}[!ht]
    \centering
    \small
    \setlength{\tabcolsep}{8pt} 
    \caption{\textbf{Confusion matrix on POPE-Popular.} Compared to the baseline, CAST significantly reduces the number of ``No'' samples incorrectly predicted as ``Yes'' (from 274 to 120), effectively mitigating the ``Yes Bias''.}    
    \begin{tabular}{lcccc}
        \toprule
        \multirow{2}{*}{} & \multicolumn{2}{c}{\textbf{Baseline}} & \multicolumn{2}{c}{\textbf{CAST}} \\
        \cmidrule(lr){2-3} \cmidrule(lr){4-5}
         & \textbf{Pred: yes} & \textbf{Pred: no} & \textbf{Pred: yes} & \textbf{Pred: no} \\
        \midrule
        \textbf{Golden: yes} & 1360 & 140  & 1277 & 223 \\
        \textbf{Golden: no}  & 274  & 1226 & 120  & 1380 \\ 
        \bottomrule
    \end{tabular}
    \label{tab:cm_popular}
\end{table}

\begin{table}[!ht]
    \centering
    
    \small
    \setlength{\tabcolsep}{8pt}
    \caption{\textbf{Confusion matrix on POPE-Random.} Similarly, on the random split, CAST drops the false positive rate drastically (from 197 to 83), effectively mitigating the ``Yes Bias''.}    
    \begin{tabular}{lcccc}
        \toprule
        \multirow{2}{*}{} & \multicolumn{2}{c}{\textbf{Baseline}} & \multicolumn{2}{c}{\textbf{CAST}} \\
        \cmidrule(lr){2-3} \cmidrule(lr){4-5}
         & \textbf{Pred: yes} & \textbf{Pred: no} & \textbf{Pred: yes} & \textbf{Pred: no} \\
        \midrule
        \textbf{Golden: yes} & 1340 & 160  & 1290 & 210 \\
        \textbf{Golden: no}  & 197  & 1303 & 83   & 1417 \\ 
        \bottomrule
    \end{tabular}
    \label{tab:cm_random}
\end{table}

\section{Case Study for Caption Queries}
\label{sec:case_cap}

\begin{figure}[!ht]
  \includegraphics[width=1.0\linewidth]{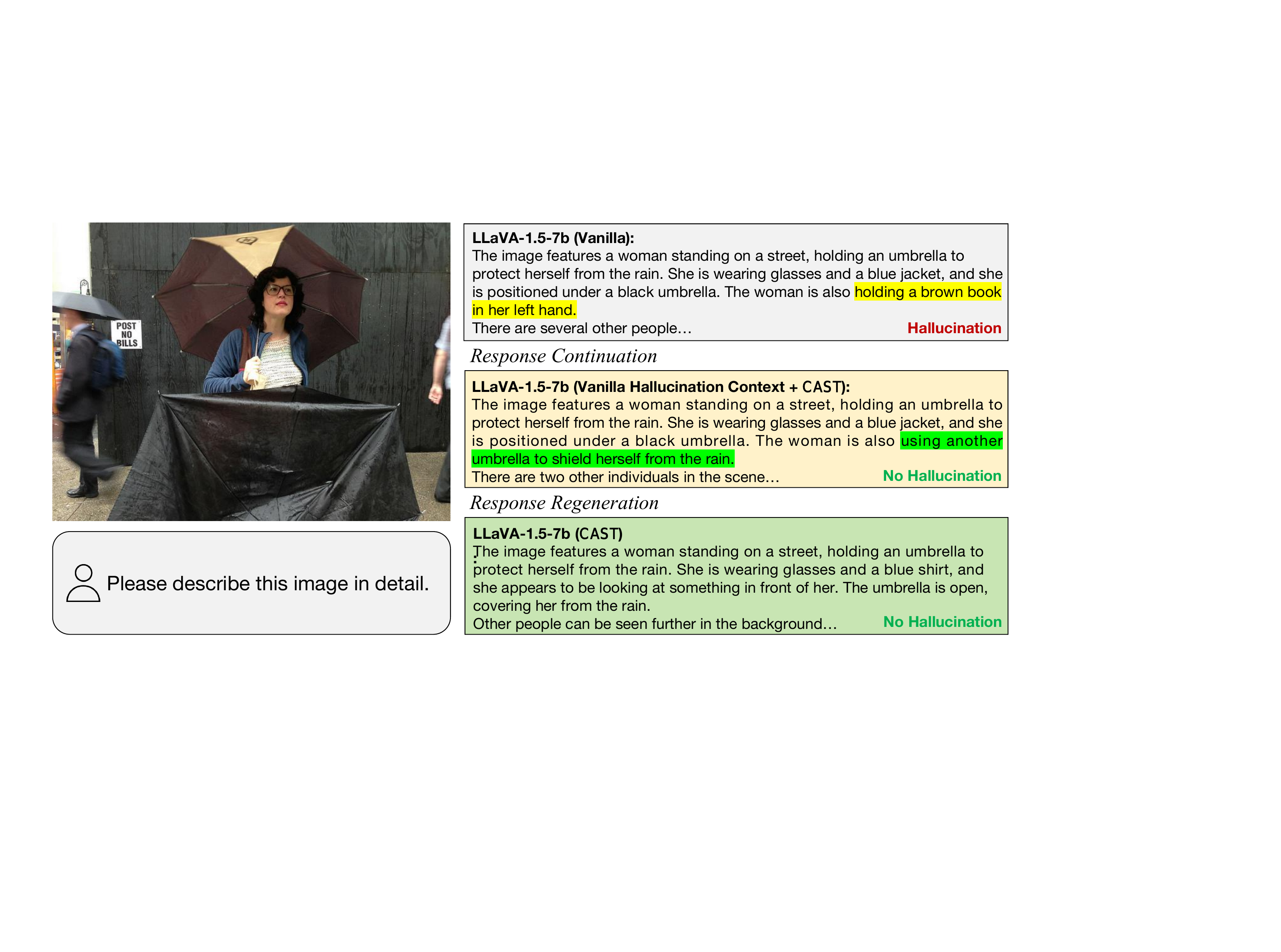}
  \caption{Case study of caption task on CHAIR. }
  \label{fig:case_study}
\end{figure}
CAST remains effective in caption task, which is attributed to the enhancement in visual attention. As shown in Figure \ref{fig:case_study}, CAST effectively mitigates object hallucination not only during the regeneration of new responses, but also when extending hallucinated contexts, highlighting its fine-grained, token-level object hallucination mitigation capability.

\section{Case Study for Non-caption Queries}
\label{sec:H}
More case studies when fed non-caption queries are shown as follows.

\begin{figure*}[!ht]
\centering
  \includegraphics[width=0.8\linewidth]{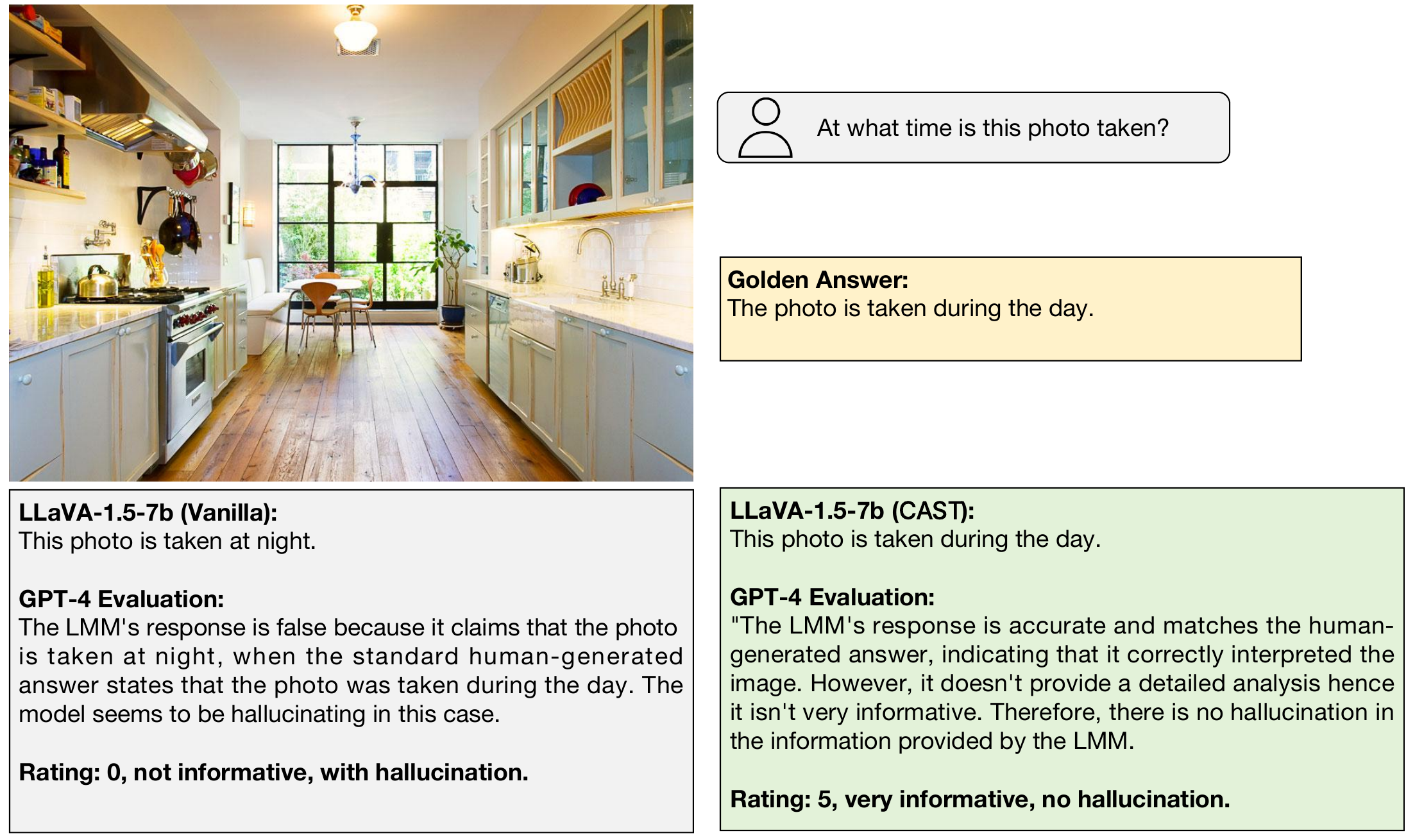}
  \caption{Non-caption query case of LLaVA-1.5-7b on MMHal-Bench.}
  \label{fig:case5}
\end{figure*}

\begin{figure*}[!ht]
\centering
  \includegraphics[width=0.9\linewidth]{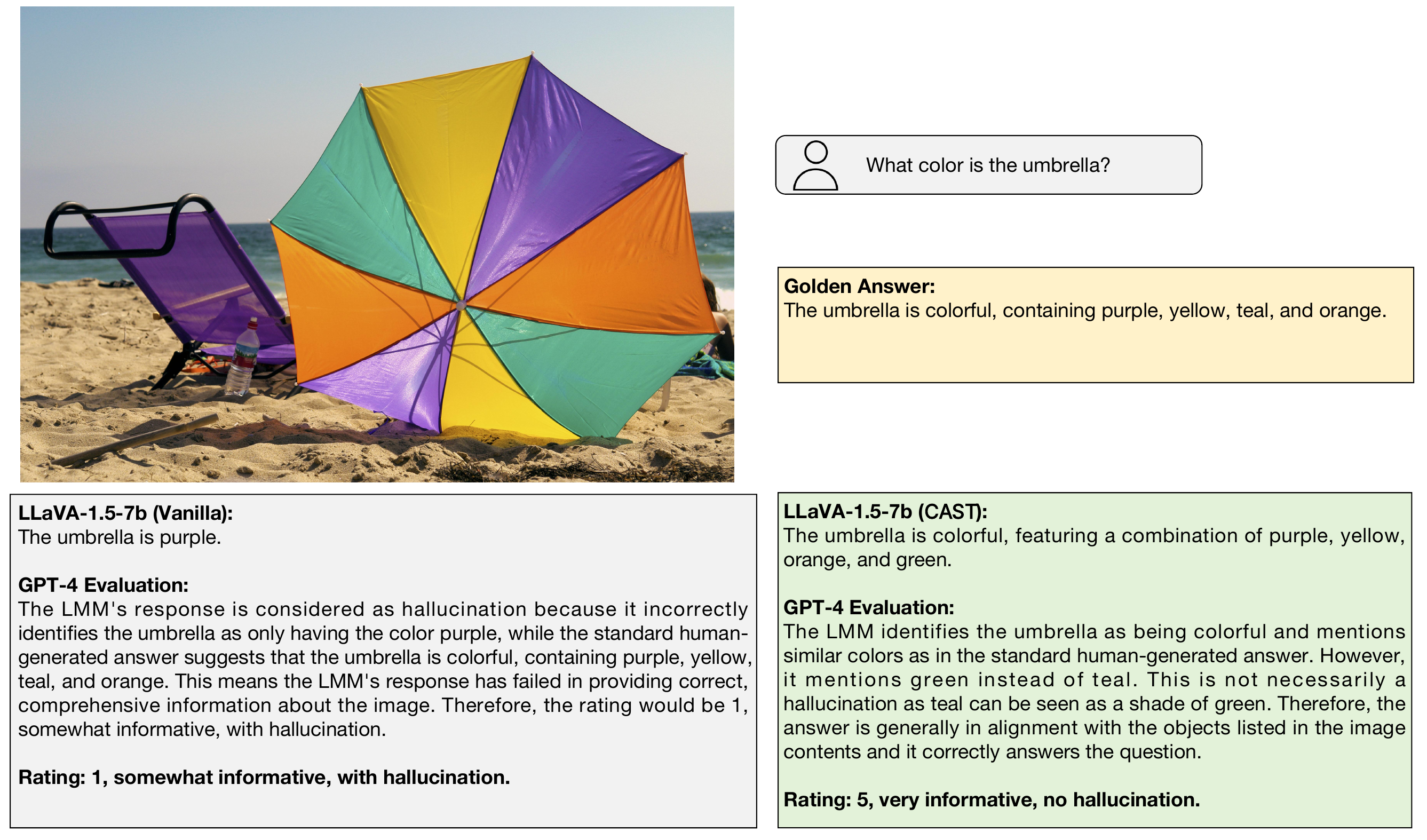}
  \caption{Non-caption query case of LLaVA-1.5-7b on MMHal-Bench.}
  \label{fig:case6}
\end{figure*}

\section{Usage of Large Language Models}
\label{sec:llm_usage}
\subsection{Assistance for Writing Polishing}
During the writing process, we employed GPT-4o \citep{hurst2024gpt} for writing polishing. In particular, we utilized LLM assistance in the method section to articulate more clearly the motivation, implementation, and corresponding mathematical formulations of the CAST approach. In addition, we applied moderate polishing to the abstract and introduction to further enhance the readability and academic rigor of the paper.
\subsection{Assistance for Benchmark Evaluation}
In conducting experiments with MMHal-Bench, we employed GPT-4 \citep{achiam2023gpt} as an evaluation tool to assess hallucination capabilities.


\end{document}